%% file: main.tex
\pdfoutput=1
\documentclass[10pt, logo, twocolumn, copyright]{nvidiatechreport}

\usepackage{caption}
\usepackage[utf8]{inputenc}
\usepackage[T1]{fontenc}
\usepackage{hyperref}
\usepackage{url}

\usepackage{booktabs}
\usepackage{amsfonts}
\usepackage{nicefrac}
\usepackage{microtype}
\usepackage[dvipsnames]{xcolor}
\usepackage{multirow}
\usepackage{graphicx}
\usepackage{subcaption}
\usepackage[numbers,sort&compress]{natbib}
\usepackage{amsmath}
\usepackage{comment}
\usepackage{makecell}
\usepackage[capitalize]{cleveref}

\crefname{section}{Sec.}{Secs.}
\Crefname{section}{Section}{Sections}
\Crefname{table}{Table}{Tables}
\crefname{table}{Tab.}{Tabs.}

\titleformat{\section}
  {\large\bfseries\headingfont}{}{0em}{#1}
\titleformat{\subsection}
  {\bfseries}{}{0em}{#1}
\titleformat{\subsubsection}
  {\bfseries\itshape}{}{0em}{#1}


\definecolor{alitblue}{RGB}{0,102,204}

\title{Accelerating Video Diffusion via Training-Free Trajectory Routing}

\author{Mustafa Munir, Huy Vu, Shreyas Misra, Rohit Jena,
Sajad Norouzi, Ali Taghibakhshi\textsuperscript{*}, Anis Ahmad, Anjul Patney,
Pavlo Molchanov, Nima Tajbakhsh\\
{\Affilfont \textsuperscript{*}Project Lead}}

\begin{abstract}
\textbf{Abstract:}
Video diffusion is computationally expensive, as it requires executing a large model across many denoising steps. Even with step-distillation, inference remains expensive because every distilled step still requires a costly model evaluation. We present \textbf{TRACK}: TRajectory-Aware Capacity routing via top-K selection, a heterogeneous denoising strategy that switches between compatible large and small models at selected steps, reducing the average cost per denoising evaluation. The switching steps are determined using a calibration process. TRACK first rolls out a reference trajectory with the large model. Then at each step, the small model's prediction is also collected and compared against the large model's prediction to obtain a relative disagreement score. Both models receive the same latent, timestep, conditioning, and guidance inputs. Aggregating this signal over a calibration set produces a disagreement score map across diffusion steps, which determines a switching policy for an efficient inference process: quality-sensitive steps keep using the large model, while steps with low disagreement scores are routed to the small model. Inference executes only the selected model at each step, requiring no retraining, architecture or scheduler changes, or online dual-model evaluation. Across Wan 2.1, Cosmos 3, TurboDiffusion, and FastVideo, TRACK yields $1.95\times$, $2.04\times$--$2.73\times$, $2.69\times$, and $2.17\times$ speedups, respectively, with comparable aggregate quality and high diversity retention. TRACK thereby establishes automated, training-free model switching as a practical acceleration paradigm for video diffusion.
\end{abstract}

\begin{document}
\maketitle

\begin{figure*}[t]
  \centering
  \includegraphics[width=0.91\textwidth]{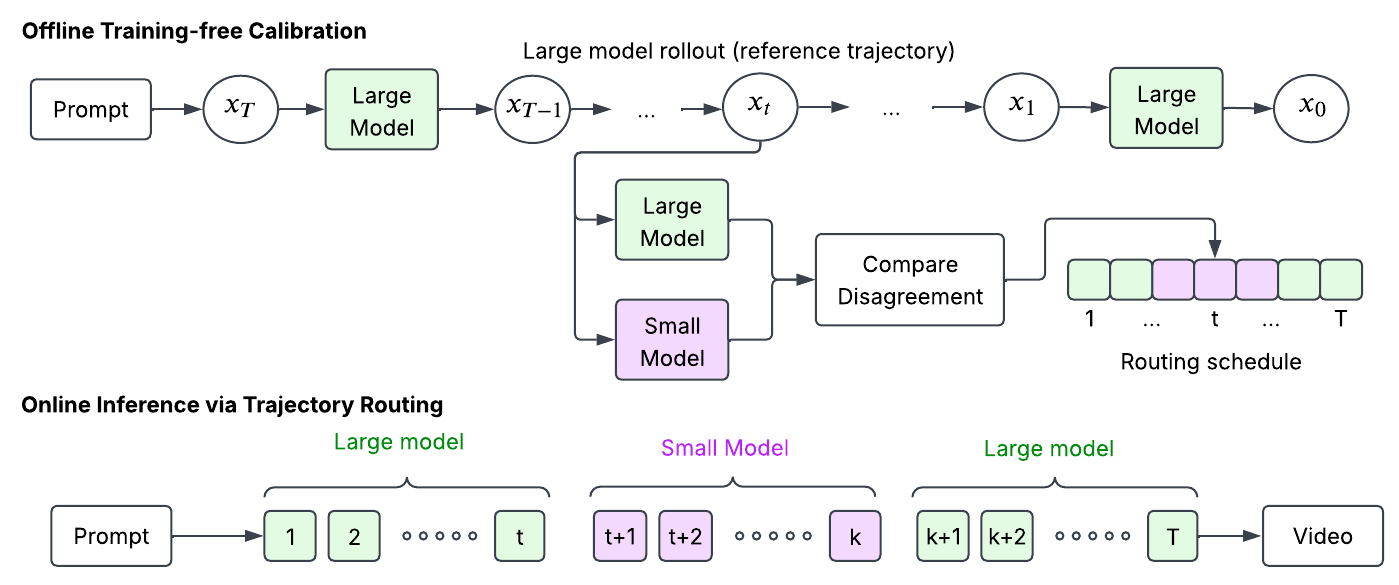}
  \caption{\textbf{Training-Free Trajectory-Aware Capacity routing.}
  \textbf{Top:} Offline calibration rolls out the all-large-model reference trajectory and, at every step, evaluates the large and small checkpoints using the same latent $x_t$, timestep, conditioning, and guidance inputs.
  Their guided-prediction disagreement is aggregated across calibration prompts to produce a switching policy that minimizes disagreement scores.
  \textbf{Bottom:} Online inference applies that policy while preserving the shared latent representation and scheduler update: quality-sensitive steps keep using the large model (green), while low disagreement steps use the small model (purple), and only one denoiser runs at each step.
  The method requires no retraining, architecture changes, scheduler changes, or online dual-model evaluation.}
  \label{fig:tfems_overview}
\end{figure*}


\section{Introduction}
\label{sec:intro}

Video diffusion models synthesize high-fidelity visual content with realistic motion, but inference latency remains a primary bottleneck for practical deployment~\cite{ho2020denoising, dhariwal2021diffusion, rombach2022high, gupta2023photorealisticvideogeneration, esser2023structure}. Advanced systems such as Wan~2.1~\cite{wan2025} and Cosmos~3~\cite{nvidia2026cosmos} achieve high physical realism and temporal consistency, yet sampling requires repeatedly executing multi-billion-parameter models over high-dimensional spatiotemporal latents. Model families often provide smaller, lightweight checkpoints that run significantly faster, though at the cost of lower generation quality. This raises a natural deployment question: \emph{can we selectively combine the speed of the small model with the quality of the large one?}

Most diffusion acceleration methods reduce the number of denoising evaluations through distillation or specialized few-step objectives~\cite{yin2023onestepdistribution,SD_Turbo,song2024sdxsrealtimeonesteplatent}, or reduce the cost within each evaluation through caching, streaming, model-sharing, and sparse computation~\cite{bolya2023tokenmergingfaststable, kodaira2025streamdiffusion, munir2026smoothdiffusion, zhang2025fastvideo, zhang2025turbodiffusion, pan2025tstitch}. These directions are essential, but they leave a complementary source of redundancy unexploited. After a sampler and step count have been selected, the remaining evaluations are still assigned to one model at uniform cost. The model switching approach instead reduces the average model cost per remaining step. It can therefore multiply the gains from step reduction rather than compete with them.

Allocating heterogeneous model capacity across video diffusion trajectories introduces distinct efficiency and quality challenges. Naive capacity allocation (i.e., model switching) based on rigid or heuristic temporal boundaries or simple monotonic handoffs fails to capture the complex spatiotemporal dynamics of video generation. For example, two models can exhibit similar aggregate prediction errors while disagreeing on spatial detail, camera motion, or frame-to-frame consistency at specific timesteps. Furthermore, uncalibrated capacity reduction risks altering structural commitment and collapsing sample diversity across seeds. Therefore, a practical efficiency framework must identify model-switchability at step-level for a given checkpoint pair. This helps dynamically protect quality-sensitive steps with a large model while offloading switchable steps to the smaller model.

We introduce \textbf{TRACK}: TRajectory-Aware Capacity routing via top-K selection, an automated framework that turns the trajectory capacity allocation problem into an offline calibration problem. Given independently trained large and small checkpoints sharing a latent space and scheduler (e.g., large 14B and small 1.3B checkpoints of the Wan model family), TRACK rolls out an all-large reference trajectory on a calibration set. At each step, both denoisers are evaluated on the exact same reference latent $x_t$, timestep, conditioning, and guidance. We then measure their normalized relative disagreement score to determine steps where the small model accurately approximates the large model. Aggregating these scores produces a disagreement map, which leads to an offline switching policy that routes quality-sensitive steps to the large model and less sensitive, switchable steps to the small model. During inference, exactly one denoiser runs at each step, requiring no retraining, architecture or scheduler modifications, or online dual-model evaluations.

We evaluate TRACK across four major video diffusion pipelines spanning both many-step and step-distilled models. TRACK achieves $1.95\times$ speedup on Wan~2.1 and  $2.04\times$--$2.73\times$ on Cosmos~3 while maintaining visual quality. On few-step distilled pipelines, TRACK yields $2.17\times$ speedup on three-step FastVideo and $2.69\times$ on four-step TurboDiffusion, showing that our model switching policy is effective with aggressive step reduction to maximize inference speed. Beyond latency reduction gains, our spatial and latent-temporal frequency analyses demonstrate that prediction disagreement is phase-dependent, explaining intuitively why intermediate timesteps safely tolerate reduced model capacity. Finally, multi-seed evaluations confirm that calibrated, disagreement-guided switching preserves the diversity across generated outputs. TRACK thus establishes training-free model switching as a practical acceleration paradigm that optimizes inference efficiency while maintaining quality and diversity. We summarize our contributions as follows:
\begin{itemize}
    \item We introduce TRACK, a training-free heterogeneous denoising strategy that generalizes prior large--small video diffusion switching from a single permanent handoff to disagreement-calibrated, fixed-budget allocation at individual denoising steps, enabling contiguous and non-contiguous schedules without modifying weights, architectures, or schedulers.
    \item We validate TRACK across four video diffusion families spanning many-step and distilled few-step pipelines, obtaining $1.95\times$--$2.73\times$ speedups at comparable visual quality and similar diversity retention as the all-large model baseline.
    \item We provide spatial and latent-temporal frequency analyses that explain the phase-dependent substitutability of large and small denoisers.
\end{itemize}


\section{Related Work}
\label{sec:related_work}

\subsection{Diffusion Models for Visual Generation}

Denoising diffusion models learn a reverse process that transforms noise into samples through a sequence of denoising updates~\cite{ho2020denoising, song2021denoising}.
Improved training and guidance made diffusion competitive for high-fidelity conditional image synthesis~\cite{dhariwal2021diffusion,ho2022classifier}, while latent diffusion reduced computation by moving the reverse process from pixel space to a learned latent representation~\cite{rombach2022high}.
Video diffusion extends this framework with temporal modeling so that appearance, structure, and motion remain coherent across frames~\cite{gupta2023photorealisticvideogeneration,esser2023structure}.
Recent model families such as Wan~2.1 and Cosmos~3 demonstrate the quality and breadth of modern video generation systems~\cite{wan2025,nvidia2026cosmos, blattmann2023stable}.
However, their high-dimensional spatiotemporal latents and heavy multi-billion parameter denoisers (e.g., Wan 14B, Cosmos 3 - Super 64B) make repeated sampling evaluations a primary deployment efficiency bottleneck.

\subsection{Efficient Diffusion Inference}
  
Diffusion acceleration generally acts on either the total number of denoising steps or the cost of each individual step. Step-reduction methods use distillation or specialized training objectives to compress sampling trajectories. Faster ODE solvers such as DPM-Solver~\cite{lu2022dpm} reduce the number of required steps without retraining, while distillation-based methods including Distribution Matching Distillation, SD-Turbo, SDXS, and LCM push further toward one- or few-step generation~\cite{yin2023onestepdistribution, SD_Turbo, song2024sdxsrealtimeonesteplatent, luo2023latent}.
While these methods reduce latency, they require additional training and can introduce model- and budget-dependent quality tradeoffs.

Other techniques optimize individual forward passes through spatial redundancy reduction, feature caching, and streaming execution~\cite{bolya2023tokenmergingfaststable, ma2024deepcache, zhao2024pab, kodaira2025streamdiffusion, liang2025looking, munir2026smoothdiffusion}.
FastVideo and TurboDiffusion combine few-step sampling with video-specific model and system optimizations~\cite{zhang2025fastvideo, zhang2025turbodiffusion}.
TRACK is complementary to both categories: it retains the target sampler and step count while reducing the average model evaluation cost across the denoising process. Its application to three-step FastVideo and four-step TurboDiffusion directly demonstrates this complementarity with step reduction.

\subsection{Adaptive and Multi-Model Diffusion} 
The diffusion denoising process exhibits distinct generation characteristics across timesteps. Spectral analyses show that deep networks process spatial frequency bands non-uniformly~\cite{hyvarinen2009natural, rahaman2019spectral, wang2020highfrequency, park2022vision, munir2025data}, and diffusion trajectories generally progress from coarse structural formation toward finer detail refinement~\cite{pan2025tstitch}. This timestep dependence has motivated multi-expert and adaptive-computation approaches that allocate specialized denoisers, frequency components, or subnetworks to different portions of the sampling trajectory~\cite{balaji2022ediff, yang2023diffusion, taghibakhshi2026star}. Such methods, however, typically require specialized training, architectural modifications, auxiliary control mechanisms, or additional expert parameters. Related ideas arise in model stitching, where representations from networks of different capacities are connected to obtain flexible efficiency--accuracy trade-offs~\cite{pan2023stitchable, pan2024snnetv2}. T-Stitch applies model switching to image diffusion by executing a single, one-way handoff—using a small model for the initial steps before switching to a large model for the remainder of the trajectory~\cite{pan2025tstitch}.

Other methods like SRDiffusion~\cite{cheng2025srdiffusion} accelerate video diffusion on Wan~\cite{wan2025} and CogVideoX~\cite{yang2024cogvideox} by switching from a large to a small model based on a runtime latent-change signal computed from the large model alone, enforcing a permanent one-way handoff. This approach has two limitations. First, it prevents the large model from resuming even when disagreement rises again, which our frequency analysis in Figure \ref{fig:tfems_error_analysis} shows occurs in the final denoising steps. Second, since the switching criterion is based only on the large model's predictions, it does not take into account the small model's predictions; therefore, it cannot directly measure how well the small model approximates the large one at a given step. TRACK addresses both limitations: it computes cross-model disagreement on shared reference latents offline, and derives a flexible policy to switch back and forth between models when needed.

Other methods like~\citep{jena2025elucidating} anneal between the score functions of a base model and its reward-fine-tuned counterpart over the denoising trajectory to preserve complementary characteristics of the two models. TRACK differs in both how its routing policy is obtained and the setting in which it is applied. Rather than using prescribed temporal schedules or learning adapters, experts, or control modules, TRACK evaluates compatible, independently trained large and small checkpoints on identical reference latents and uses the disagreement scores from their predictions to derive an offline, per-step switching policy. At inference, only the selected denoiser is evaluated at each step, requiring no retraining, architectural modification, or online dual-model evaluation.

While prior work has demonstrated training-free large--small switching for video diffusion via a single permanent handoff, TRACK generalizes this to disagreement-calibrated, fixed-budget allocation at individual denoising steps, enabling contiguous and non-contiguous schedules and compatibility with step-distilled pipelines.


\section{Proposed Methodology}
\label{sec:method}

TRACK accelerates video diffusion by applying a precomputed step-dependent model-capacity schedule without modifying weights or altering schedulers.
Figure~\ref{fig:tfems_overview} illustrates the overall approach.
The method operates in two stages: an offline calibration phase that measures step-level model disagreement on a shared reference trajectory and an execution phase that evaluates exactly one denoiser per step according to a calibrated switching policy. The following section~\ref{subsec:preliminaries} formalizes the compatible sampling framework, while section~\ref{subsec:selection_policy} details reference-trajectory calibration and derives the disagreement-based switching policy.

\subsection{Preliminaries}
\label{subsec:preliminaries}

\paragraph{Video diffusion sampling.}
Let $x_{t_i}\in\mathbb{R}^{C\times F\times H\times W}$ denote a noisy video latent at denoising-step index $i$, where $C$, $F$, $H$, and $W$ denote the channel, frame, height, and width dimensions, respectively.
The sampler follows a decreasing sequence of scheduler timesteps $t_0>t_1>\cdots>t_{N-1}$.

Given a large model $L$ and a small model $S$, along with prompt conditioning $\tau$, negative prompt $\emptyset$, and guidance scale $w$, the denoiser $m\in\{L,S\}$ yields the guided prediction (with classifier-free guidance):

\begin{equation}
    \begin{split}
        p_m(x_{t_i},t_i) &= f_m(x_{t_i},t_i,\emptyset) \\
        &\quad + w \big(f_m(x_{t_i},t_i,\tau) - f_m(x_{t_i},t_i,\emptyset)\big),
    \end{split}
    \label{eq:cfg_prediction}
\end{equation}

The scheduler maps this prediction to the next latent:
\begin{equation}
    x_{t_{i+1}}
    = \mathcal{S}\!\left(x_{t_i},p_m(x_{t_i},t_i),t_i,h_i\right),
    \label{eq:scheduler_update}
\end{equation}
where $h_i$ denotes any scheduler history required by a multistep solver.
TRACK preserves the original timestep sequence, prediction target, scheduler, and scheduler-state evolution; it alters only which denoiser $m\in\{L,S\}$ supplies $p_m(x_{t_i},t_i)$ in Eq.~\eqref{eq:scheduler_update}.

\paragraph{Compatible model pairs.}
TRACK operates on pairs of large and small checkpoints that can be seamlessly interchanged within the same sampling trajectory. They must share the latent representation, output shape, conditioning interfaces, prediction target, and scheduler semantics.
Although their internal architectures and parameter counts may differ, their guided predictions defined in Eq.~\eqref{eq:cfg_prediction} must be semantically identical for the scheduler, ensuring either checkpoint can validly advance the shared latent state in Eq.~\eqref{eq:scheduler_update}. In practice, this applies to diffusion model families which have multiple model sizes but are based on a similar training and inference process.

\paragraph{Offline Per-Step Capacity Scheduling.}
An offline-derived switching policy $\pi:\{0,\ldots,N-1\}\rightarrow\{L,S\}$ assigns a model size to each denoising-step index.
Inference then follows
\begin{equation}
    x_{t_{i+1}}
    = \mathcal{S}\!\left(
        x_{t_i},
        p_{\pi(i)}(x_{t_i},t_i),
        t_i,
        h_i
      \right).
    \label{eq:elastic_sampling}
\end{equation}
With $C_m$ denoting the wall-clock latency of a single evaluation of model $m$ on the target hardware, the total denoising cost of the policy is approximately $\sum_i C_{\pi(i)}$, excluding shared text encoding, decoding, and scheduler overhead. The objective is therefore to minimize this total denoising cost by deriving a policy $\pi$ that selectively assigns the small checkpoint to switchable steps while retaining the large checkpoint at quality-sensitive steps. TRACK derives this $\pi$ offline from a small one-time calibration set, then inference executes directly from the calibrated offline policy with no further online overhead.

\subsection{Disagreement-Based  Switching Policy}
\label{subsec:selection_policy}

\subsubsection{Reference-Trajectory Calibration}
\label{subsec:reference_calibration}

Let $\mathcal{D}=\{d_j\}_{j=1}^{M}$ be a small calibration set.
For each prompt $d_j$, we initialize the pipeline normally and roll out the complete trajectory using only the large checkpoint.
Immediately before each scheduler update, we evaluate both small and large checkpoints on the same reference latent $x_{t_k}^{(j)}$, timestep $t_k$, conditioning, negative conditioning, and guidance scale.
The paired predictions therefore differ only in the denoiser that produced them. After recording the pair, the large-model prediction advances the trajectory, ensuring that every subsequent probe remains on the all-large reference path.

This procedure produces one paired large--small prediction at every denoising step and for every calibration prompt.
It does not modify either checkpoint or the underlying sampler.
Although both denoisers are evaluated during calibration, this cost is incurred only offline and is not part of the online inference.

\begin{table*}[t]
  \centering
  \small
  \setlength{\tabcolsep}{4.5pt}
  \resizebox{\textwidth}{!}{%
  \begin{tabular}{lccccccc}
    \toprule
    Model & Configuration & \makecell{Switched /\\Total Steps} & Speedup & Flicker $\uparrow$ & Motion $\uparrow$ & Subject $\uparrow$ & Background $\uparrow$ \\
    \midrule
    \multirow{2}{*}{\shortstack[l]{Wan 2.1\\\cite{wan2025}}}
                   & All-Large Model (14B) & 0/50  & $1.00\times$ & $96.71$ & $98.31$ & $94.62$ & $94.79$ \\
                   & TRACK            & 30/50 & $1.95\times$ & $96.86$ & $98.39$ & $94.81$ & $94.99$  \\
    \midrule
    \multirow{3}{*}{\shortstack[l]{Cosmos 3\\\cite{nvidia2026cosmos}}}
                   & All-Large Model (Super)  & 0/35  & $1.00\times$ & $98.96$ & $99.41$ & $96.77$ & $95.39$  \\
                   & TRACK (Super/Nano)  & 24/35 & $2.04\times$ & $99.06$ & $99.46$ & $96.93$ & $95.55$  \\
                   & TRACK (Super/Edge)  & 24/35 & $2.73\times$ & $98.33$ & $98.98$ & $97.24$ & $95.04$ \\
    \midrule
    \multirow{2}{*}{\shortstack[l]{TurboDiffusion\\\cite{zhang2025turbodiffusion}}}
                   & All-Large Model (14B) & 0/4  & $1.00\times$ & $97.26$ & $98.64$ & $94.10$ & $93.32$ \\
                   & TRACK            & 3/4  & $2.69\times$ & $96.94$ & $98.42$ & $94.10$ & $93.48$ \\
    \midrule
    \multirow{2}{*}{\shortstack[l]{FastVideo\\\cite{zhang2025fastvideo}}}
                   & All-Large Model (14B) & 0/3  & $1.00\times$ & $98.19$ & $99.15$ & $95.81$ & $95.91$ \\
                   & TRACK            & 2/3  & $2.17\times$ & $97.99$ & $99.12$ & $96.07$ & $96.16$ \\
    \bottomrule
  \end{tabular}%
  }
  \caption{\textbf{Speed--quality tradeoffs across video pipelines.}
  Comparison between All-Large (baseline), and the TRACK method (ours) across full model pipelines (Cosmos 3 \cite{nvidia2026cosmos} and Wan 2.1 \cite{wan2025}) and step-distilled pipelines (TurboDiffusion's Wan 2.1 \cite{zhang2025turbodiffusion}, FastVideo's Wan 2.1 \cite{zhang2025fastvideo}). Performance is evaluated on the same 250 prompts for all configurations using VBench's temporal flicker, motion smoothness, subject consistency, and background consistency~\cite{huang2023vbench}.Our method shows significant speed-up at comparable aggregate quality.
  \label{tab:cross_family_results}}
\end{table*}

\subsubsection{Normalized Prediction Disagreement}
For calibration prompt $j$ and denoising step $k$, both the small and large model are evaluated on the same all-large reference latent $x_{t_k}^{(j)}$, yielding $p_S(x_{t_k}^{(j)},t_k)$ and $p_L(x_{t_k}^{(j)},t_k)$, respectively.
We first compute the normalized guided-prediction disagreement score. Here, the score is normalized by the prediction magnitude of the large model:
\begin{equation}
    r_{j,k}
    = \frac{\left\|p_S(x_{t_k}^{(j)},t_k)-p_L(x_{t_k}^{(j)},t_k)\right\|_2}
    {\max\!\left(\left\|p_L(x_{t_k}^{(j)},t_k)\right\|_2,\epsilon\right)}
    \label{eq:raw_relative_disagreement}
\end{equation}

where $t_k$ is the scheduler timestep at denoising-step index $k$, and $\epsilon > 0$ is a small constant for numerical stability that prevents division by zero when the large-model prediction norm is negligible. Then, TRACK computes the mean normalized score across $M$ calibration prompts as the final disagreement score:
\begin{equation}
    q_k
    = \frac{1}{M}\sum_{j=1}^{M} r_{j,k},
    \label{eq:tfems_score}
\end{equation}

Lower $q_k$ indicates closer large--small agreement and therefore a safer candidate for using small-model execution instead of large-model.

\subsubsection{Constrained Trajectory Divergence Minimization using Top-K Switching}

The calibration score provides a non-negative trajectory divergence surrogate using the relative difference between the guided predictions of the small and large models. 
Let $W_k = q_k^2$ be a squared divergence cost associated with choosing the smaller model, and let $\beta_k$ denote a relaxed switching variable between the small model $S$ and large model $L$ for denoising iteration $k$. We associate a benefit of \(-C_0\sum_k\beta_k\) with using the small model, where \(C_0\) is a user-specified constant representing a cost we want to minimize (e.g. latency on specific hardware). This gives us a box-constrained optimization problem:
 
\begin{equation}
    \arg\min_{\beta_0,\ldots,\beta_{N-1} \in [0,1]}\sum_k (W_k - C_0)\beta_k 
    \label{eq:copt}
\end{equation}

where $N$ is the total number of denoising steps.

Although \cref{eq:copt} is a continuous optimization problem, its objective is linear and separable in $\beta_k$. Therefore, for $W_k\neq C_0$, an optimal solution is
\begin{equation}
  \beta_k^\star =
  \begin{cases}
      1, & W_k<C_0,\\
      0, & W_k>C_0.
  \end{cases}
  \label{eq:threshold_solution}
\end{equation}
When $W_k=C_0$, any value in $[0,1]$ is optimal; in particular, an
integral solution can always be selected.

For a fixed switching budget $K$, TRACK routes the $K$ steps with the lowest calibrated disagreement scores $q_k$ to the small model and retains the large model at the remaining steps. Since $q_k\geq 0$, ranking by $q_k$ is equivalent to ranking by $W_k=q_k^2$, and choosing $C_0$ between the $K$-th and $(K{+}1)$-th smallest $W_k$ makes the threshold solution in \cref{eq:threshold_solution} select exactly these $K$ steps; specifying $C_0$ is therefore equivalent to specifying $K$. This special case of the continuous relaxation thus admits a discrete optimal solution without a separate mixed-integer formulation, and provides direct control over the compute budget by specifying the number, or equivalently the fraction, of denoising steps assigned to the small model.

\begin{figure}[t]
  \centering
  \includegraphics[width=1.0\columnwidth]{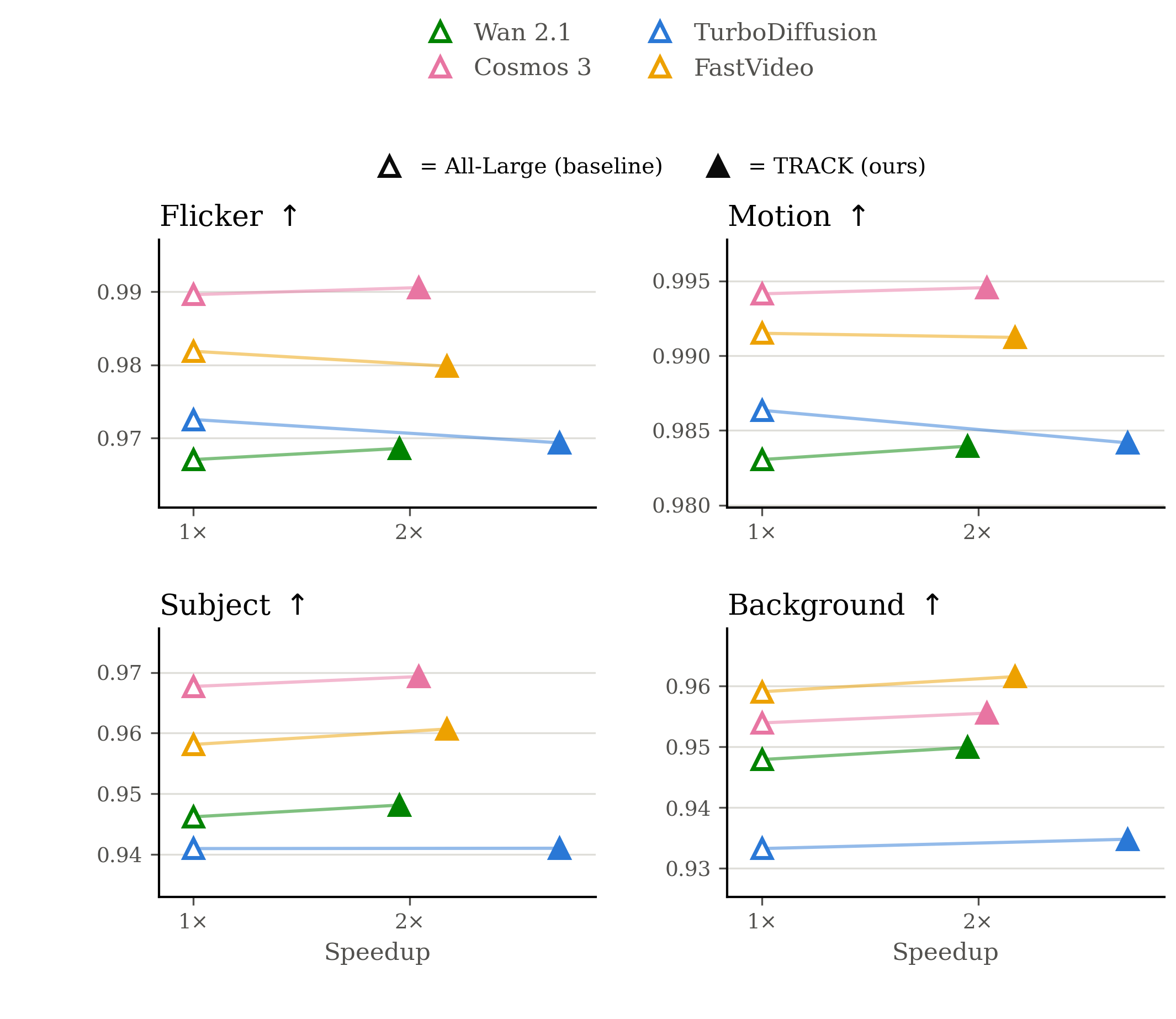}
  \caption{\textbf{Speed--quality tradeoffs across video pipelines.}
  Each panel compares the all-large model (baseline) and TRACK (ours) switching policy. All models retain comparable aggregate quality at the selected operating points in most quality dimensions.
  \label{fig:cross_family_pareto}}
\end{figure}

\subsection{Inference}
\label{subsec:elastic_inference}

During inference, generation strictly follows the offline pre-computed policy $\pi$ via Eq.~\eqref{eq:elastic_sampling}. The pipeline evaluates exactly one checkpoint per step, seamlessly passing the shared latent, conditioning, and scheduler state between the two models. Because the step allocation is determined entirely offline, inference requires no learned routing modules or simultaneous dual-model evaluations.
Assuming a single-pass latency of $C_m$ for checkpoint $m$, the total denoising cost reduces directly to $\sum_i C_{\pi(i)}$, plus the standard scheduler overhead. Because policy calibration is a one-time offline setup for a given sampler configuration, it introduces zero overhead during actual deployment.

\begin{figure}[t]
  \centering
  \includegraphics[width=0.9\columnwidth]{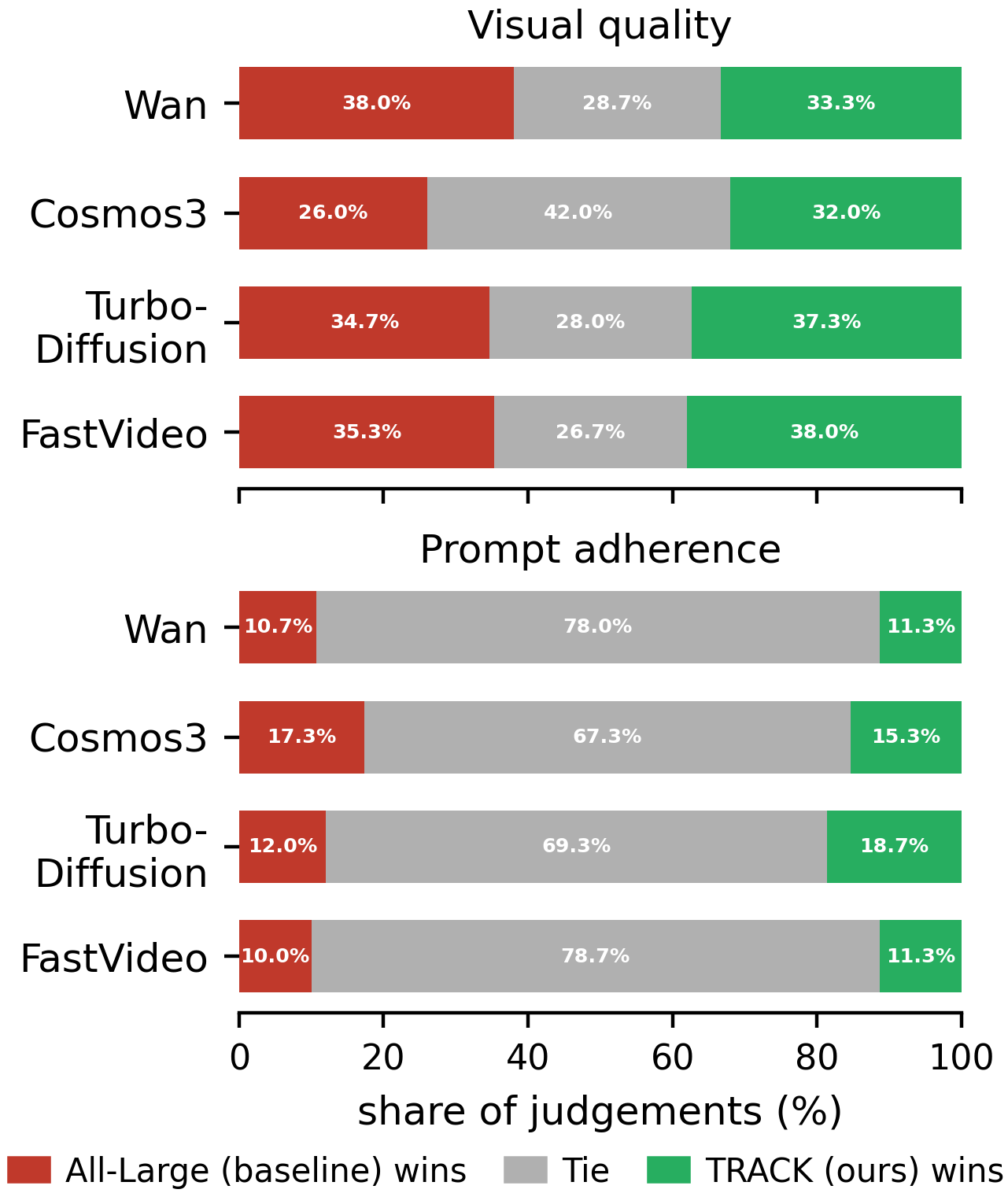}
  \caption{\textbf{Human evaluation study.}
  Human evaluators compare TRACK outputs against the corresponding all-large model across Wan~2.1, Cosmos~3, TurboDiffusion, and FastVideo.
  Each horizontal bar reports the percentage of comparisons rated as TRACK \emph{Win}, \emph{Tie}, or \emph{Lose} against the all-large baseline.}
  \label{fig:user_study}
\end{figure}

\section{Experimental Results}
\label{sec:experiments}

\subsection{Experimental Setup}
\label{subsec:experimental_setup}

\paragraph{Pipelines and Switching Budgets.} We evaluate TRACK on four text-to-video pipelines using their native inference pipelines. We pair the large and small checkpoints of Wan~2.1 (14B/1.3B, 50 steps)~\cite{wan2025}, Cosmos~3 (Super/Nano and Super/Edge, 35 steps)~\cite{nvidia2026cosmos}, FastVideo (14B/1.3B, 3 steps)~\cite{zhang2025fastvideo}, and TurboDiffusion (14B/1.3B, 4 steps)~\cite{zhang2025turbodiffusion}. All pipelines generate 81-frame, $480\times832$ videos. Through an ablation study (Appendix~\ref{subsec:budget_ablation}), we select the operating-point switching budget $K$ as the maximum number of small-model steps that maintains quality comparable to the all-large baseline.

\vspace{-2mm}
\paragraph{Evaluation Protocols.} We evaluate speed and quality across 250 text-video prompt pairs from VBench~\cite{huang2023vbench}, EvalCrafter \cite{Liu_evalcrafter_2024_CVPR}, T2V-CompBench \cite{Sun_t2v_compbench_2025_CVPR}, and manually designed prompts. For quality and temporal consistency, we report VBench's temporal flicker, motion smoothness, subject consistency (DINO ViT-B/16~\cite{caron2021emerging}), and background consistency. Generative diversity is measured via mean pairwise DreamSim distance~\cite{fu2023dreamsim} over ten prompts with eight shared seeds per policy.

\begin{figure}[t]
  \centering
  \includegraphics[width=\columnwidth]{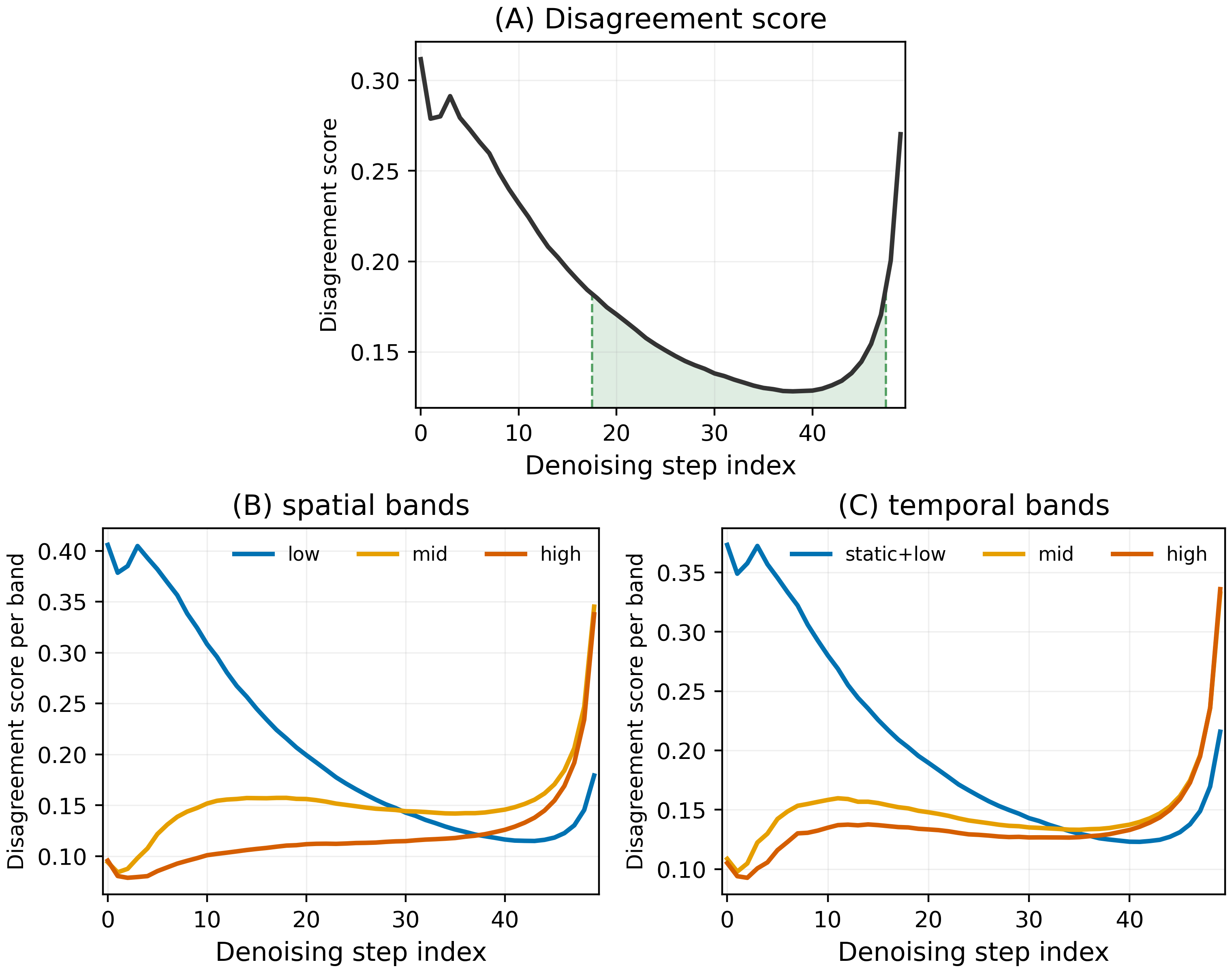}
  \caption{\textbf{Analysis of Wan 2.1's disagreement scores across denoising steps.} Disagreement scores along the denoising trajectory for Wan 2.1 (14B vs. 1.3B), averaged across 150 prompts at 50 steps. (A) Scalar disagreement score. The green lines denote top-K steps with lowest disagreement score suitable for switching. (B, C) The same score decomposed over spatial and temporal frequency bands.}
  \label{fig:tfems_error_analysis}
\end{figure}

\vspace{-2mm}
\paragraph{Human Evaluation.} We conducted a pairwise human evaluation comparing TRACK switching against the all-large baseline across all four pipelines. Evaluators were shown side-by-side videos generated from the same prompt and asked to judge which was better on two axes: visual quality and prompt adherence. We collected a minimum of 150 judgments per pipeline (30 prompts $\times$ 5 raters each).

\subsection{Quantitative and Qualitative Results}
\label{subsec:main_results}

Table~\ref{tab:cross_family_results}, together with
Figure~\ref{fig:cross_family_pareto}, summarizes the selected
operating point for each pipeline. TRACK accelerates all four families, spanning 50-, 35-, four-, and three-step trajectories, without changing their native schedulers or step counts.
On the 50-step Wan~2.1 pipeline, the calibrated switching policy achieves a $1.95\times$ speedup.
For Cosmos~3, the policy yields a $2.04\times$--$2.73\times$ speedup depending on whether Nano or Edge model is chosen for the smaller model.
TRACK is also effective with step-distilled pipelines, achieving a $2.17\times$ speedup on 3-step FastVideo and a $2.69\times$ speedup on 4-step TurboDiffusion.

In terms of quality, Wan~2.1 preserves subject and background consistency relative to the all-large model. Cosmos~3 retains subject and background consistency as well, while TurboDiffusion and FastVideo are similarly comparable across all metrics, demonstrating that our model switching approach remains effective after few-step distillation.

These results show that TRACK reduces the cost of the denoising evaluations on both conventional many-step sampling and aggressive step distillation.
Crucially, offline calibration on a small prompt set generalizes seamlessly to the full evaluation benchmark without manual step tuning. This demonstrates that a disagreement-based switching policy provides a stable, prompt-independent criterion for automated offline model allocation across both many-step and step-distilled video diffusion systems.

\subsection{User Study}
\label{subsec:User_Study}
\begin{figure*}[t]
  \centering
  \includegraphics[width=\textwidth]{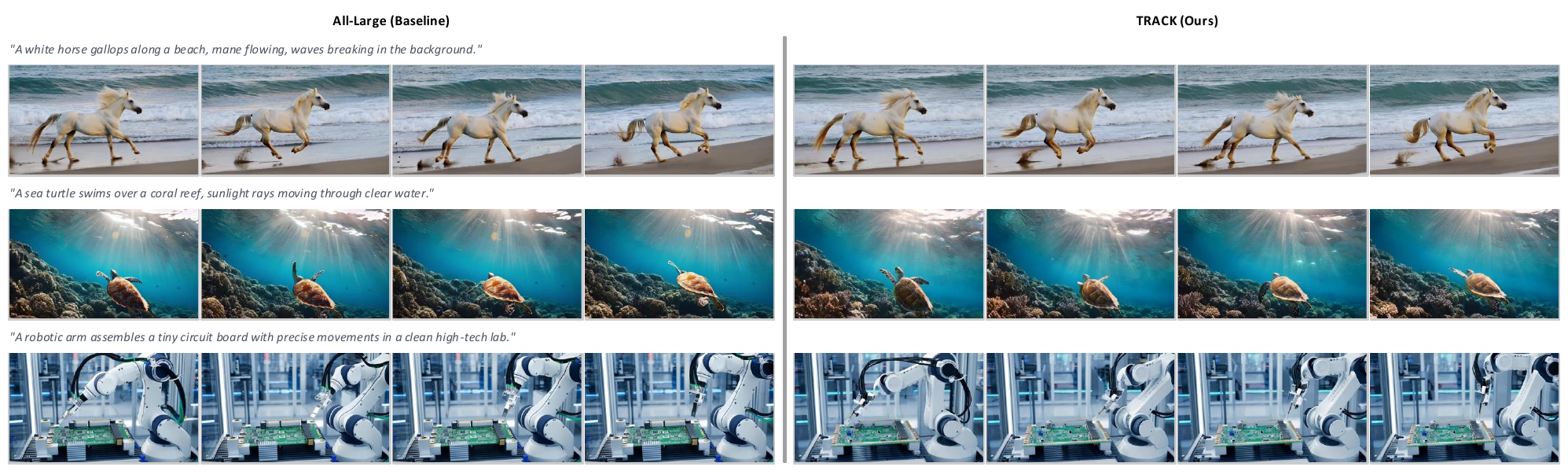}
  \caption{\textbf{Qualitative comparison.}
  Comparing frames from videos generated with TRACK switching policy and the all-large 14B model for Wan 2.1. Our approach preserves the scene composition, subject identity, and motion pattern of the large-model outputs.}
  \label{fig:qualitative_turbo}
\end{figure*}

Figure~\ref{fig:user_study} summarizes the results of the human evaluation study. On prompt adherence, evaluators found the two pipelines indistinguishable in the majority of comparisons across all models (67--79\% tie rate), with win and lose rates near parity. On visual quality, judgments were more evenly distributed, but the TRACK switching approach matched or edged the full model: TRACK wins are comparable or slightly exceeded full-model wins on TurboDiffusion (37.7\% vs.\ 34.7\%), FastVideo (38\% vs.\ 35.3\%), Cosmos 3 (32.0\% vs.\ 26.0\%), and Wan~2.1 (33.3\% vs.\ 38.0\%), with Cosmos~3 showing the largest tie share (42.0\%). Across both dimensions, no model family showed a consistent preference for the full model, confirming that the speedups reported in Table~\ref{tab:cross_family_results} come without a perceptible quality penalty. Figure \ref{fig:qualitative_turbo} illustrates qualitatively the videos generated by our method and baseline. Further qualitative results can be found in Figure \ref{fig:qualitative_appendix} in the Appendix.

\begin{table*}[t]
    \centering
    \scriptsize
    \resizebox{\textwidth}{!}{%
    \begin{tabular}{lccccccc}
      \toprule
      Policy &
      Flicker $\uparrow$ &
      Motion $\uparrow$ &
      Subject $\uparrow$ &
      Background $\uparrow$ &
      DreamSim Diversity Retention $\uparrow$ \\
      \midrule
      All-Large
      & $96.30$
      & $98.08$
      & $94.28$
      & $94.95$
      & $100.0\%$ \\ 
      All-Small
      & $96.35$ ($+0.05\%$)
      & $97.63$ ($-0.46\%$)
      & $93.83$ ($-0.48\%$)
      & $94.17$ ($-0.82\%$)
      & $79.4\%$ ($-20.6\%$) \\[-4pt]
      \multicolumn{8}{l}{\color{gray}\rule{\linewidth}{0.3pt}} \\
      First-$K$
      & $95.39$ ($-0.95\%$)
      & $96.90$ ($-1.20\%$)
      & $93.77$ ($-0.54\%$)
      & $94.14$ ($-0.85\%$)
      & $84.6\%$ ($-15.4\%$) \\
      Last-$K$
      & $96.04$ ($-0.27\%$)
      & $97.89$ ($-0.19\%$)
      & $94.05$ ($-0.24\%$)
      & $94.51$ ($-0.46\%$)
      & $97.8\%$ ($-2.2\%$) \\
      Random-$K$
      & $95.95$ ($-0.36\%$)
      & $97.55$ ($-0.54\%$)
      & $93.91$ ($-0.39\%$)
      & $94.49$ ($-0.48\%$)
      & $85.7\%$ ($-14.3\%$) \\
      TRACK Top-$K$
      & \textbf{96.47 (+0.18\%)}
      & \textbf{98.14 (+0.06\%)}
      & \textbf{94.41 (+0.14\%)}
      & \textbf{94.88 (-0.07\%)}
      & \textbf{98.3\% (-1.7\%)} \\
      \bottomrule
    \end{tabular}%
    }
  \caption{\textbf{Equal-budget switching policy comparison.} We compare TRACK Top-$K$ policy against heuristic policies (First-$K$, Last-$K$, and Random-$K$) on Wan 2.1 with $K=30$, with each policy replacing exactly $K$ denoising steps with the small model. 
  All switching policies therefore use the same number of large- and small-model evaluations.
  Temporal flicker, motion smoothness, subject consistency, and background consistency are evaluated on a 150-prompt evaluation set. DreamSim diversity \cite{fu2023dreamsim} is evaluated separately using ten prompts with eight shared seeds per policy. TRACK Top-$K$ provides the best performance with the same switching budget.}
  \label{tab:equal_budget_comparison}
\end{table*}

\subsection{Analysis of Disagreement across Denoising Steps}
\label{subsec:frequency_analysis}

  We visualize the disagreement scores across diffusion steps to understand at which steps the small model can replace the large one. Figure~\ref{fig:tfems_error_analysis}A shows the
  disagreement curve for Wan 2.1 over a 50-step denoising trajectory, averaged across 150 prompts. The error is high at the initial and final steps and low in between, with the minimum falling in the latter half of the
  trajectory. We observe the same pattern for Cosmos 3 (Figure~\ref{fig:tfems_summary_paper_appendix} in Appendix). For these models, switching to the small model is therefore safest in the middle steps.
  
  To understand what drives this landscape, we decompose the disagreement scores over spatial and temporal frequency bands (low, mid, high) using an orthonormal Fourier transform.
   Figures~\ref{fig:tfems_error_analysis}B and~\ref{fig:tfems_error_analysis}C show that the early-step disagreement is dominated by low-frequency components, whereas the late-step
  error is dominated by mid and high frequencies. This is consistent with the coarse-to-fine nature of the denoising process: early steps commit to global structure, so any
  disagreement between the two models appears in the low-frequency band, while late steps refine fine-grained detail, shifting the disagreement to higher frequencies. The disagreement
  is lowest in the middle of the trajectory - mostly on the latter half, where the coarse structure is already fixed and fine detail has not yet been resolved. Hence, these steps are the most promising for switchability without causing significant disagreement between the large and small models. Appendix reports additional disagreement landscapes and frequency analyses for Cosmos 3 \cite{nvidia2026cosmos}, TurboDiffusion \cite{zhang2025turbodiffusion}, and FastVideo \cite{zhang2025fastvideo}.

\subsection{Equal-Budget Policy Ablations and Diversity Retention Analysis}
\label{subsec:equal_budget_diversity}
To isolate the effect of where model switching occurs from the overall amount of compute reduction, we compare TRACK against three equal-budget switching baselines. This ablation helps us understand how our policy compares to other naive or heuristic approaches at a pre-determined number of switching steps. Specifically, each policy replaces exactly $K$ out of the $N$ denoising steps with the small model: First-$K$ replaces the initial $K$ steps, Last-$K$ replaces the final $K$ steps, and Random-$K$ randomly selects $K$ steps. In contrast, TRACK Top-$K$ selects the $K$ steps with the lowest calibrated disagreement scores $q_k$. Here, First-$K$ is aligned with the T-stitch \cite{pan2025tstitch} method, in which the first steps use the small model then switch entirely to the large model for the rest.
On the other hand, Last-$K$ is conceptually similar to SRDiffusion's \cite{cheng2025srdiffusion} approach, in which the first steps use large model while using small model for the rest. One nuanced difference is, Last-$K$ fixes the number of switching steps while SRDiffusion dynamically determines the number switching steps based on input prompts. We also conduct additional experiments comparing our method and SRDiffusion \cite{cheng2025srdiffusion} in the Appendix.

All policies use identical prompts and total numbers of large- and small-model evaluations. Besides quality comparison, we also evaluate generative sample diversity without conflating it with semantic or temporal drift. We measure mean pairwise DreamSim distance \cite{fu2023dreamsim} over eight seeds, four frames per video, and ten prompts. Following established diversity protocols~\cite{gandikota2026distilling}, we adapt five broad prompt categories from their benchmark and supplement them with three motion-heavy prompts and two control prompts.

Table~\ref{tab:equal_budget_comparison} summarizes quality and diversity metrics across these switching policies on Wan 2.1 ($K=30/50$). We find that in both quality and diversity retention metrics, our approach outperforms all switching policies compared. We outperform First-K by a particularly large margin in diversity retention, likely because generation diversity primarily stems from the large model's predictions in the initial steps \cite{gandikota2026distilling}, which First-K has replaced with the small model.
Compared to Last-K (and additionally SRDiffusion in the Appendix Section \ref{sec:supp_srdiffusion_comparison}), our method also outperforms across quality and diversity metrics. This might be explained by our analysis in section \ref{subsec:frequency_analysis}, which shows that the last diffusion steps contain high disagreement between large and small model when finalizing the videos' high-frequency details - hence, the pipeline should switch back to the large model. This observation once again illustrates our approach's flexibility when switching models across steps, compared to other methods' one-time handoffs—either small-to-large or large-to-small. Table \ref{tab:equal_budget_comparison} confirms the effectiveness of the switching policy built on evidence-based calibration analysis over heuristic policies.

Regarding diversity retention, since in our approach, the switchable steps are mostly in the middle diffusion steps, all initial steps use the large model. This might explain why our method maintains high diversity retention compared to other methods, confirming the findings in ~\cite{gandikota2026distilling}. Further representative multi-seed visual generations, as well as additional evaluations following the diversity evaluation protocol from~\cite{gandikota2026distilling}, are reported in Section~\ref{sec:supp_diversity_extended} of the Appendix.

\section{Conclusion}
\label{sec:conclusion}

We introduced TRACK, a training-free approach for converting compatible, independently trained large and small video diffusion checkpoints into a switching denoising system.
TRACK calibrates the checkpoints on the same all-large reference latents, uses a normalized guided-prediction disagreement score to identify switchable denoising steps, and executes a switching policy with exactly one model evaluation per step.
It therefore changes neither model weights nor scheduler behavior and introduces no online dual-model comparison.

Across Wan~2.1, Cosmos~3, FastVideo, and TurboDiffusion, TRACK provides $1.95\times$--$2.73\times$ speedups at selected comparable-quality operating points.
The gains on three- and four-step pipelines show that our method complements step-distillation by further lowering the average cost of the distilled steps.
Spatial and latent-temporal analyses further show that large--small disagreement is structured across the trajectory, providing an empirical and intuitive explanation for model-specific step selection. We further tested our approach against other heuristic switching policies with the same step budget and demonstrated that our method outperforms these baselines in both quality and diversity metrics. Together, these results establish TRACK's model-switching policy as a practical additional axis for accelerating video diffusion while maintaining quality.

{
    \small
    \bibliographystyle{ieeenat_fullname}
    \bibliography{refs}
}

\input{sec/suppl}

\end{document}

%% file: sec/suppl.tex
\let\titleold\title
\def\maketitlesupplementary
   {
   \newpage
       \twocolumn[
        \centering
        \Large
        \textbf{Accelerating Video Diffusion via Training-Free Trajectory Routing}\\
        \vspace{0.5em}Supplementary Material \\
        \vspace{1.0em}
       ] 
   }

\clearpage
\pagenumbering{roman}
\maketitlesupplementary
\appendix

\section{Ablation on Determining the Switching Budget ($K$)}
\label{subsec:budget_ablation}

To understand the tradeoff between compute reduction and generation quality, which helps determine the optimal switching budget, we conduct an ablation study varying the switching budget $K$ (the number of denoising steps routed to the small model) across all four pipelines. For this analysis, we report the average of temporal flicker, motion smoothness, subject consistency, and background consistency scores, providing a unified metric for visual fidelity. The switching policies are generated using the TRACK Top-$K$ selection criterion derived in Section~\ref{sec:method}, meaning steps are routed in order of lowest normalized disagreement score.

\begin{figure}[ht]
  \centering
  \includegraphics[width=\columnwidth]{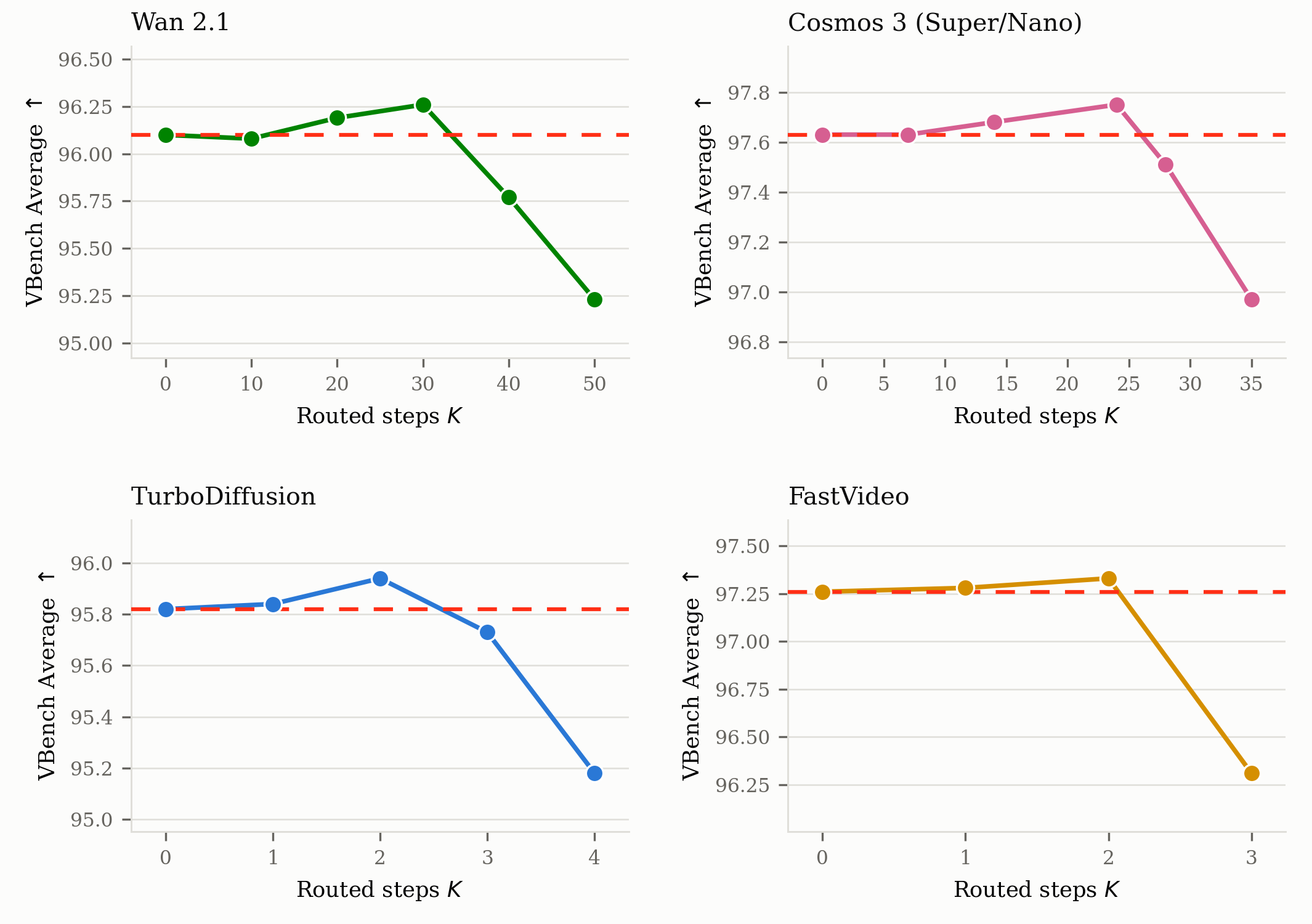}
  \caption{\textbf{Effect of trajectory routed budget steps ($K$) on visual quality.} We evaluate the average of the temporal flicker, motion smoothness, subject consistency, and background consistency scores from VBench across varying numbers of small-model steps ($K$). 
  From this study, we can determine the maximum $K$ used in the TRACK algorithm before the quality starts to degrade.}
  \label{fig:budget_sweep_consistency}
\end{figure}

Figure~\ref{fig:budget_sweep_consistency} reports the visual consistency across varying budgets. The results demonstrate a clear ``plateau of switchability'' for video diffusion models. As $K$ increases from $0$ (the all-large baseline), visual consistency initially remains stable and can even improve. This confirms that a substantial portion of the denoising trajectory simply does not require the representational capacity of the largest network. 

However, as $K$ approaches the total number of sampling steps $N$, the capacity reduction begins to encroach on quality-sensitive steps, causing a sharp degradation in visual fidelity. For instance, increasing the budget to $K=28$ on Cosmos~3 or $K=40$ on Wan 2.1 severely penalizes consistency.

\section{Extended Analysis of Disagreement Scores Across Denoising Steps}
\label{sec:supp_frequency}

\begin{figure}[t]
  \centering
  \includegraphics[width=\columnwidth]{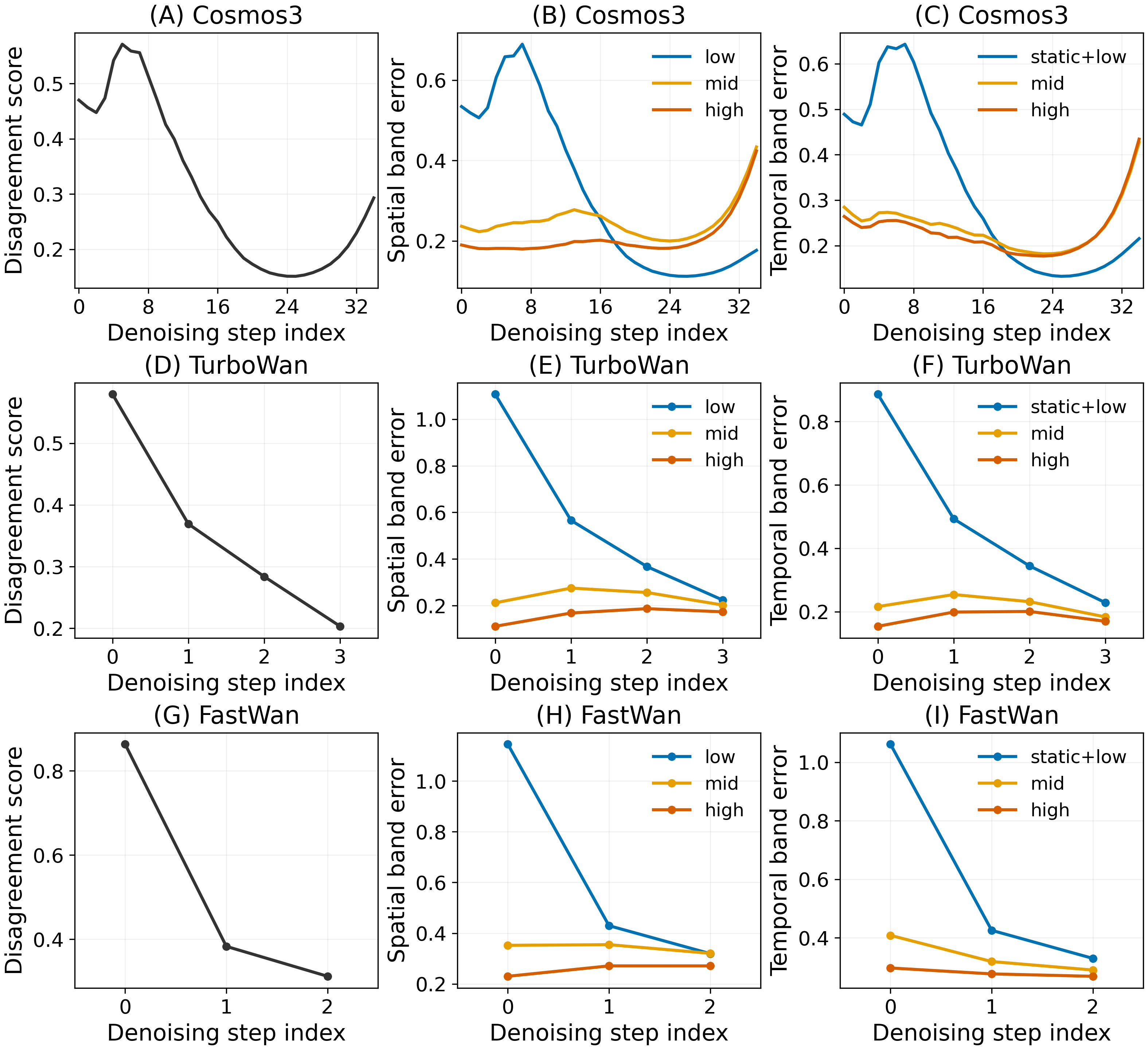}
  \caption{\textbf{TRACK disagreement score across denoising steps analysis across models.}
  Disagreement score error for Cosmos 3 (A--C, 35 steps), TurboDiffusion (D--F, 4 steps), and FastVideo (G--I, 3 steps), averaged across prompts. Columns in order of: scalar disagreement score; spatial-band frequencies amplitude; temporal-band frequencies amplitude.}
  \label{fig:tfems_summary_paper_appendix}
\end{figure}

Figures~\ref{fig:tfems_summary_paper_appendix}A–C show that Cosmos 3 closely follows the behavior of Wan 2.1. The disagreement score is high at the initial and final steps and reaches its minimum in the latter half of the trajectory. The band decomposition reproduces the same frequency handoff: early-step disagreement is concentrated in the low band, while the late-step rise is carried by the mid and high bands. The U-shaped disagreement landscape is therefore not specific to Wan 2.1, and switching is again best placed in the middle steps. TurboDiffusion \cite{zhang2025turbodiffusion} and FastVideo \cite{zhang2025fastvideo} show a different landscape (Figures ~\ref{fig:tfems_summary_paper_appendix}D–I). The score is largest at the first step and decreases monotonically until the last, with no late-step rise. The band decomposition shows that the disagreement remains low-frequency dominated across all steps — the low band carries the majority of the error energy at every step.

\section{Energy Consumption and Efficiency Gains}
\label{subsec:energy_savings}

Beyond accelerating inference latency, TRACK's inference efficiency can be translated into substantial reductions in GPU energy consumption. This is a critical metric for the deployment of generative video models at scale, where continuous execution of multi-billion parameter denoisers incurs heavy power and thermal costs.

\paragraph{Measurement Protocol.} 
We measure the hardware energy consumption of the denoising loop using the NVIDIA Management Library (NVML). For each generated video $i$, we read the cumulative GPU energy counter immediately before and after the denoising trajectory. The per-video energy consumption in joules is calculated as:
\begin{equation}
    E_i = \frac{\text{NVML}_{\mathrm{end}} - \text{NVML}_{\mathrm{start}}}{1000}.
\end{equation}
To ensure a rigorous evaluation, we compute the total percentage of energy saved as the ratio of aggregate means across the entire evaluation set, rather than the unweighted average of individual per-video percentages. Let $\bar{E}_{\text{base}}$ and $\bar{E}_{\text{TRACK}}$ denote the mean per-video energy consumption for the all-large baseline and the TRACK policy, respectively. The aggregate energy savings percentage is defined as:
\begin{equation}
    \text{Savings (\%)} = \left( 1.0 - \frac{\bar{E}_{\text{TRACK}}}{\bar{E}_{\text{base}}} \right) \times 100.
\end{equation}
All measurements are recorded on a single NVIDIA A100 GPU, isolating the denoising loop and excluding fixed overheads such as text encoding and VAE decoding.

\paragraph{Results.}
Table~\ref{tab:energy_savings} summarizes the energy reductions across all evaluated pipelines. TRACK cuts the GPU energy footprint by approximately half across the board.

\begin{table}[ht]
  \centering
  \footnotesize
  \setlength{\tabcolsep}{2.5pt}
  \resizebox{\columnwidth}{!}{%
  \begin{tabular}{lcccccc}
    \toprule
    Pipeline & \makecell{Steps \\ ($K/N$)} & \makecell{Base \\ (kJ)} & \makecell{TRACK \\ (kJ)} & \makecell{Saved \\ (kJ)} & \makecell{Saved \\ (\%)} & \makecell{1K Vids \\ (kWh)} \\
    \midrule
    TurboDiff. \cite{zhang2025turbodiffusion} & 2/4   & $8.31$   & $4.52$   & $3.79$   & $45.6\%$ & $1.05$ \\
    FastVideo \cite{zhang2025fastvideo}       & 2/3   & $9.44$   & $4.30$   & $5.14$   & $54.4\%$ & $1.43$ \\
    Cosmos 3 (S/N) \cite{nvidia2026cosmos}    & 24/35 & $204.22$ & $97.07$  & $107.15$ & $52.5\%$ & $29.77$ \\
    Cosmos 3 (S/E) \cite{nvidia2026cosmos}    & 24/35 & $213.40$ & $85.93$  & $127.46$ & $59.7\%$ & $35.41$ \\
    Wan 2.1 \cite{wan2025}                    & 30/50 & $394.43$ & $199.86$ & $194.57$ & $49.3\%$ & $54.05$ \\
    \bottomrule
  \end{tabular}%
  }
   \caption{\textbf{GPU Energy Consumption and Savings.} Denoising energy is measured via NVML on a single NVIDIA A100 GPU. Percentage savings are calculated using the ratio of the aggregate means. TRACK reduces the energy footprint of video generation by roughly $45\%$ to $60\%$, saving up to $54$ kWh per 1,000 videos on many-step models and over $1$ kWh per 1,000 videos on aggressively step-distilled models. For Cosmos~3, S/N and S/E denote the Super/Nano and Super/Edge checkpoint configurations, respectively.}
  \label{tab:energy_savings}
\end{table}

\begin{table*}[ht]
    \centering
    \scriptsize
    \setlength{\tabcolsep}{3pt}
    \resizebox{\textwidth}{!}{%
    \begin{tabular}{lccccccc}
      \toprule
      Policy &
      Flicker $\uparrow$ &
      Motion $\uparrow$ &
      Subject $\uparrow$ &
      Background $\uparrow$ &
      DreamSim Diversity $\uparrow$ \\
      \midrule
      All-Large
      & $96.30$
      & $98.08$
      & $94.28$
      & \textbf{94.95}
      & \textbf{100.0\%} \\ 
      \midrule
      SRDiffusion ($0.03$) \cite{cheng2025srdiffusion}
      & $95.84$ ($-0.48\%$)
      & $97.57$ ($-0.52\%$)
      & $93.76$ ($-0.55\%$)
      & $94.35$ ($-0.63\%$)
      & $87.1\%$ ($-12.9\%$) \\
      SRDiffusion ($0.01$) \cite{cheng2025srdiffusion}
      & $95.98$ ($-0.33\%$)
      & $97.73$ ($-0.36\%$)
      & $93.94$ ($-0.36\%$)
      & $94.63$ ($-0.34\%$)
      & $92.6\%$ ($-7.4\%$) \\
      TRACK Top-$K$ (Ours)
      & \textbf{96.47 (+0.18\%)}
      & \textbf{98.14 (+0.06\%)}
      & \textbf{94.41 (+0.14\%)}
      & 94.88 (-0.07\%)
      & 98.3\% (-1.7\%) \\
      \bottomrule
    \end{tabular}%
    }
  \caption{\textbf{TRACK vs. SRDiffusion Comparison.} We compare our TRACK Top-$K$ policy ($K=30$) and two threshold variants of SRDiffusion \cite{cheng2025srdiffusion} on Wan 2.1. \textbf{Bold} indicates the best performing method.}
  \label{tab:srdiffusion_comparison}
\end{table*}

For many-step, high-capacity pipelines, the absolute energy savings are significant. On Wan 2.1 (14B to 1.3B), TRACK saves $194.57$~kJ per video, translating to $54.05$~kWh saved per 1,000 videos generated (a $49.3\%$ reduction). On Cosmos~3 (Super/Edge), TRACK reduces energy consumption by nearly $60\%$, saving $35.41$~kWh per 1,000 videos. 

Crucially, TRACK remains highly effective even on heavily distilled pipelines where the total step count is already minimized. On the three-step FastVideo pipeline and four-step TurboDiffusion pipeline, TRACK yields $54.4\%$ and $45.6\%$ energy savings, respectively. At scale, this prevents over $1$~kWh of energy waste per 1,000 videos without requiring any additional training or distillation. In practice, video diffusion pipelines generate millions of videos, and these energy savings scale directly with the number of generated videos.

\newpage

\section{Additional Quantitative Results}
\label{sec:supp_srdiffusion_comparison}

In this section, we provide a direct, isolated comparison between our proposed TRACK switching policy and SRDiffusion \cite{cheng2025srdiffusion}. SRDiffusion determines its switching point using an online, prompt-adaptive threshold based on the latent rate of change. We compare TRACK against two variants of SRDiffusion (using threshold values of $0.01$ and $0.03$). 

As shown in Table~\ref{tab:srdiffusion_comparison}, TRACK outperforms the SRDiffusion baselines across both video consistency metrics and sample diversity retention. Because SRDiffusion enforces a permanent, monotonic handoff to the small model, it struggles to preserve the high-frequency temporal details in the final denoising steps, leading to loss of DreamSim diversity. By dynamically placing the small model at optimal low-disagreement steps and returning to the large model when necessary, TRACK preserves the large model's visual fidelity and diversity.

Beyond the numerical metrics, visual inspection confirms that the structural integrity of the generated videos remains intact. As shown in Figure~\ref{fig:qualitative_turbo}, the TRACK switching policy seamlessly preserves the scene composition, subject identity, and motion patterns of the all-large baseline outputs compared to contiguous handoff methods.


\section{Additional Qualitative Results}
\label{sec:supp_qualitative_results_extended}
\begin{figure*}[t]
  \centering
  \includegraphics[width=0.95\textwidth]{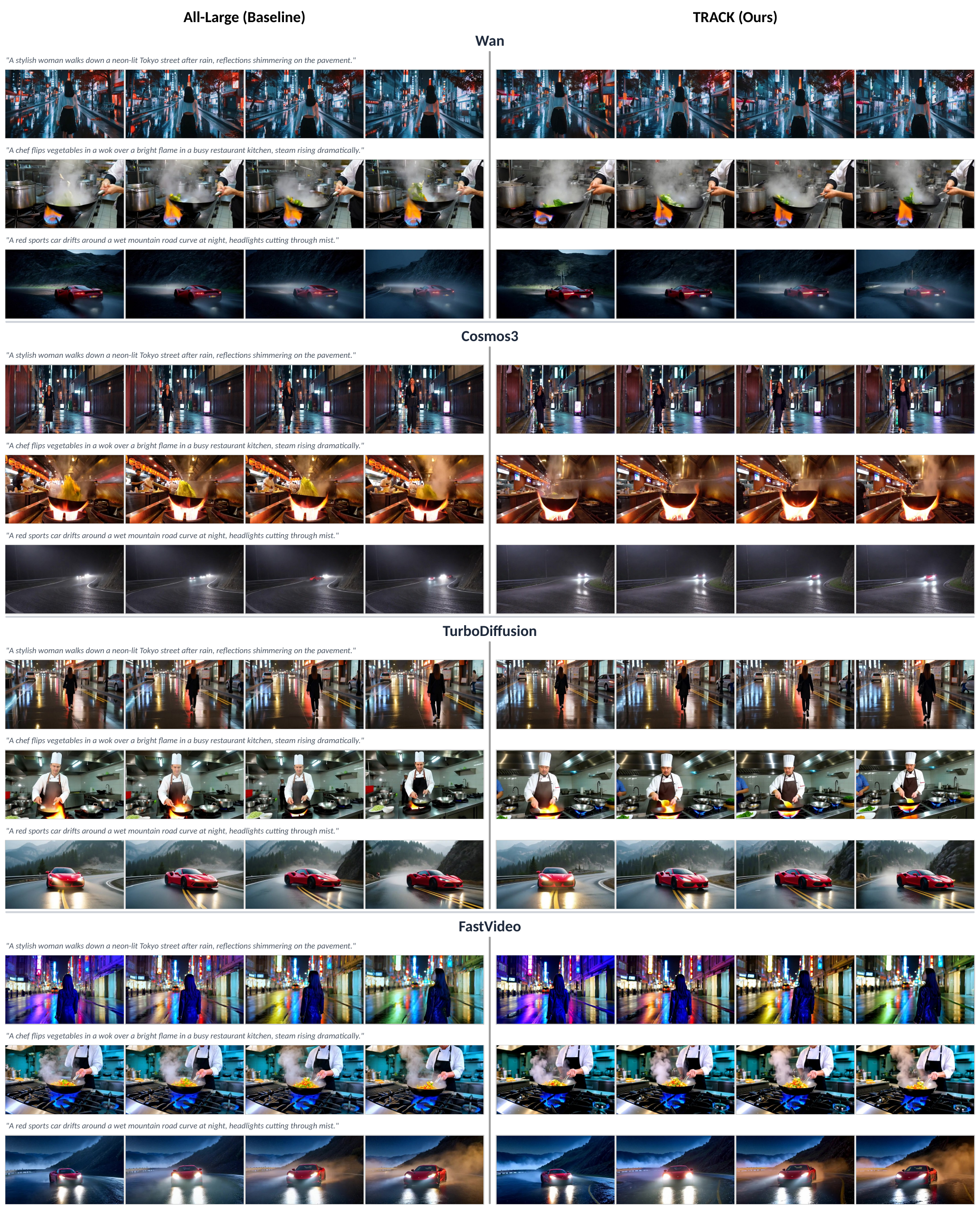}
  \caption{\textbf{Qualitative comparison across all four base models.}
  Four frames at evenly-spaced timestamps are shown for three prompts generated with each model under two configurations: the full large-model pipeline (baseline) and our TRACK pipeline (Ours).
  Across all four models: Wan 2.1, Cosmos 3, TurboDiffusion, and FastVideo, TRACK's switching policy preserves scene composition, subject identity, and motion patterns of the large-model outputs, while reducing inference latency.}
  \label{fig:qualitative_appendix}
\end{figure*}

In Figure \ref{fig:qualitative_appendix}, we provide additional side-by-side video frame comparisons across all four model families: Wan~2.1, Cosmos~3, TurboDiffusion, and FastVideo. For each model, we show frames generated from the same prompt using the all-large baseline and the TRACK switching policy. Across all families, TRACK preserves scene composition, subject identity, and motion patterns of the large-model outputs, while delivering the speedups reported in Table~\ref{tab:cross_family_results}.

\clearpage

\begin{figure*}[t!]
  \centering
  \includegraphics[width=\textwidth]{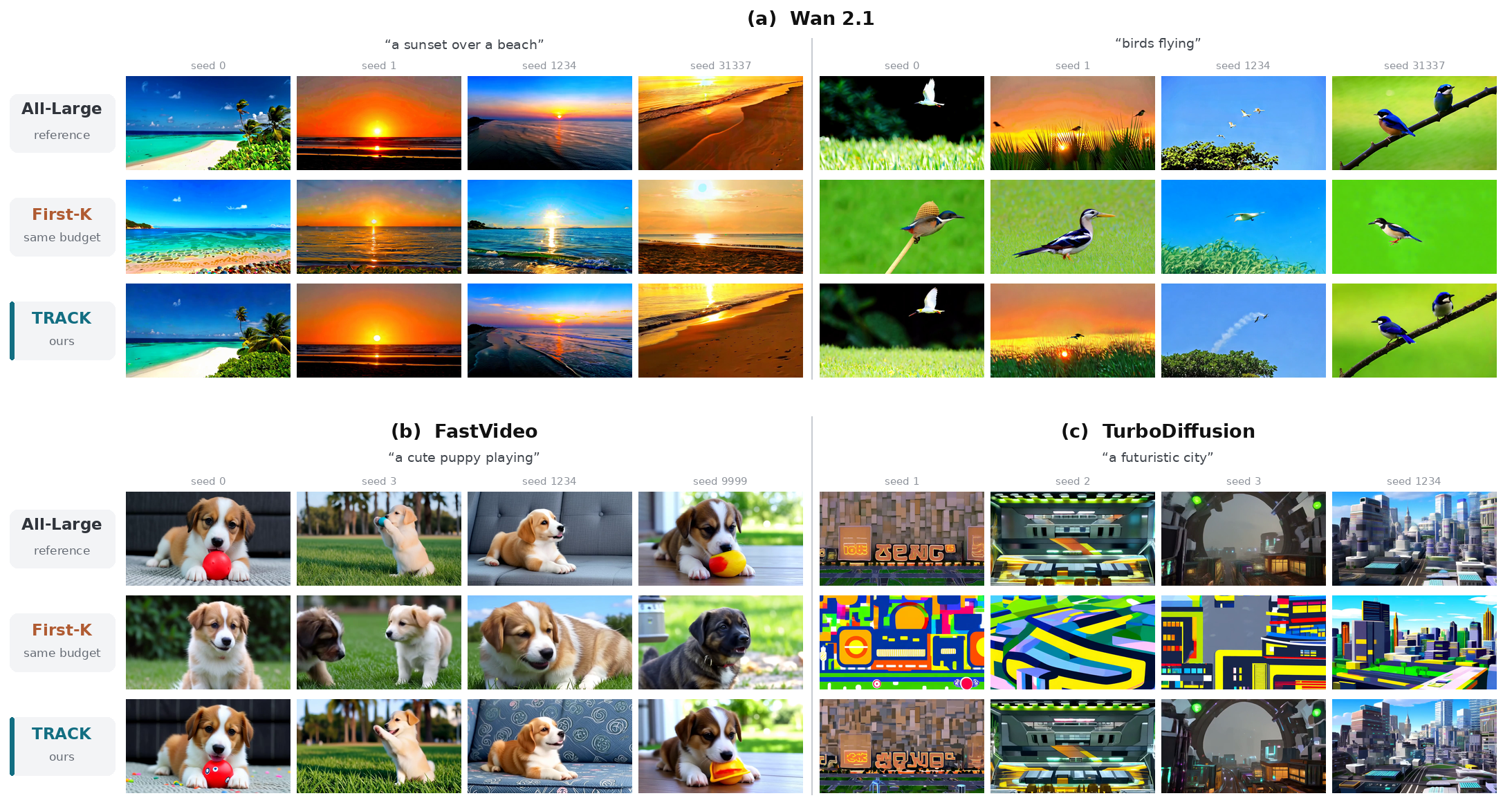}
  \caption{\textbf{Representative multi-seed generations used in the diversity analysis.}
  We show midpoint frames from four seeds per prompt, with each column using the same random seed across rows.
  Rows compare the all-large reference, a same-budget First-$K$ schedule, and TRACK; both switching policies use the same number of small-model evaluations.
  The top grid shows two prompts for Wan~2.1 with deterministic UniPC sampling \cite{zhao2023unipc}, while the lower grids show three-step FastVideo with deterministic UniPC sampling and four-step TurboDiffusion with stochastic rCM sampling. TRACK preserves the varied subjects, compositions, and styles of the all-large reference, whereas First-$K$ shifts or narrows its output modes.}
  \label{fig:supp_diversity_grids}
\end{figure*}

\section{Extended Diversity Preservation Analysis}
\label{sec:supp_diversity_extended}

When aggressively switching large-model evaluations with a smaller checkpoint, a critical concern is whether the system suffers from mode collapse or loss of sample diversity across different initial noise seeds. As shown qualitatively in Figure~\ref{fig:supp_diversity_grids}, TRACK successfully preserves the varied subjects, compositions, and styles of the all-large baseline across different random seeds, whereas naive heuristic policies like First-$K$ can severely shift or narrow the output modes. 

Following the diversity evaluation protocol established in recent literature~\cite{gandikota2026distilling}, we measure the mean pairwise DreamSim distance across generated outputs. Higher DreamSim distances indicate greater compositional and semantic diversity across seeds.

\paragraph{Evaluation Protocol and Baselines.}
We compute the DreamSim distance using five standardized prompt categories utilized in prior diffusion diversity studies \cite{gandikota2026distilling} (e.g., sunset beach, puppy, futuristic city, person, and Van Gogh art). To isolate the impact of our switching policy, we compare video generation pipelines against the image-based SDXL baseline reported in ~\cite{gandikota2026distilling}. Because video evaluation averages features across multiple frames (four frames per video across eight seeds), absolute DreamSim values differ across modalities; therefore, the primary metric of interest is the \textbf{retention percentage} relative to the all-large baseline. As reported in Table~\ref{tab:supp_diversity}, TRACK preserves $>95\%$ of the original large-model diversity across all four video pipelines, successfully mirroring the high retention dynamics observed in image-based models like SDXL \cite{podell2024sdxl}.

\begin{table}[ht]
  \centering
  \small
  \setlength{\tabcolsep}{3pt}
  \resizebox{\columnwidth}{!}{%
  \begin{tabular}{lccc}
    \toprule
    & \multicolumn{2}{c}{\textbf{Mean Pairwise DreamSim ($\uparrow$)}} & \\
    \cmidrule(lr){2-3}
    Pipeline & Base & Hybrid & Retention \\
    \midrule
    SDXL \cite{podell2024sdxl} & $0.337$ & $0.350$ & $103.9\%$ \\
    \midrule
    FastVideo \cite{zhang2025fastvideo} & $0.271$ & $0.284$ & $104.8\%$ \\
    TurboDiff. \cite{zhang2025turbodiffusion} & $0.607$ & $0.581$ & $95.7\%$ \\
    Wan 2.1 \cite{wan2025} & $0.505$ & $0.502$ & $99.4\%$ \\
    Cosmos 3 \cite{nvidia2026cosmos} & $0.550$ & $0.548$ & $99.5\%$ \\
    \bottomrule
  \end{tabular}%
  }
  \caption{\textbf{Sample Diversity Retention.} We report mean pairwise DreamSim distance on a standardized 5-category prompt set. FastVideo, TurboDiffusion, Wan 2.1, and Cosmos 3 successfully reproduce the retention dynamics observed in 2D image diffusion, preserving $> 95\%$ of large-model diversity when the highest capacity model is retained for critical denoising steps.}
  \label{tab:supp_diversity}
\end{table}

\clearpage

\section{Limitations and Future Work}
\label{sec:supp_limitations}
TRACK, similar to other diffusion model switching methods, currently still requires storing both the large and small models simultaneously, increasing deployment memory compared to one single-model pipeline. A natural direction for future work is to integrate TRACK with nested model families as introduced in the language domain~\cite{taghibakhshi2026star}, where the small denoiser is realized as a subnetwork of the large denoiser sharing the same parameters. Under such a design, the full and reduced capacity models are contained within a single checkpoint, eliminating the memory overhead of maintaining two separate models while preserving the flexibility switching benefits demonstrated here.

%% file: refs.bib
@String(CVPR= {IEEE Conf. Comput. Vis. Pattern Recog.})

@String(ICCV= {Int. Conf. Comput. Vis.})

@String(CVPRW= {IEEE Conf. Comput. Vis. Pattern Recog. Worksh.})

@String(CVPR  = {CVPR})

@String(ICCV  = {ICCV})

@String(CVPRW= {CVPRW})

@InProceedings{huang2023vbench,
     title={{VBench}: Comprehensive Benchmark Suite for Video Generative Models},
     author={Huang, Ziqi and He, Yinan and Yu, Jiashuo and Zhang, Fan and Si, Chenyang and Jiang, Yuming and Zhang, Yuanhan and Wu, Tianxing and Jin, Qingyang and Chanpaisit, Nattapol and Wang, Yaohui and Chen, Xinyuan and Wang, Limin and Lin, Dahua and Qiao, Yu and Liu, Ziwei},
     booktitle={Proceedings of the IEEE/CVF Conference on Computer Vision and Pattern Recognition},
     year={2024}
 }

@article{ho2020denoising,
  title={Denoising diffusion probabilistic models},
  author={Ho, Jonathan and Jain, Ajay and Abbeel, Pieter},
  journal={Advances in Neural Information Processing Systems},
  volume={33},
  pages={6840--6851},
  year={2020}
}

@article{dhariwal2021diffusion,
  title={Diffusion models beat gans on image synthesis},
  author={Dhariwal, Prafulla and Nichol, Alexander},
  journal={Advances in Neural Information Processing Systems},
  volume={34},
  pages={8780--8794},
  year={2021}
}

@inproceedings{rombach2022high,
  title={High-resolution image synthesis with latent diffusion models},
  author={Rombach, Robin and Blattmann, Andreas and Lorenz, Dominik and Esser, Patrick and Ommer, Bj{\"o}rn},
  booktitle={Proceedings of the IEEE/CVF Conference on Computer Vision and Pattern Recognition},
  pages={10684--10695},
  year={2022}
}

@article{ho2022classifier,
  title={Classifier-free diffusion guidance},
  author={Ho, Jonathan and Salimans, Tim},
  journal={arXiv preprint arXiv:2207.12598},
  year={2022}
}

@article{gupta2023photorealisticvideogeneration,
  title={Photorealistic video generation with diffusion models},
  author={Gupta, Agrim and Yu, Lijun and Sohn, Kihyuk and Gu, Xiuye and Hahn, Meera and Fei-Fei, Li and Essa, Irfan and Jiang, Lu and Lezama, Jos{\'e}},
  journal={arXiv preprint arXiv:2312.06662},
  year={2023}
}

@inproceedings{esser2023structure,
  title={Structure and content-guided video synthesis with diffusion models},
  author={Esser, Patrick and Chiu, Johnathan and Atighehchian, Parmida and Granskog, Jonathan and Germanidis, Anastasis},
  booktitle={Proceedings of the IEEE/CVF International Conference on Computer Vision},
  pages={7346--7356},
  year={2023}
}

@inproceedings{yin2023onestepdistribution,
  title={One-step diffusion with distribution matching distillation},
  author={Yin, Tianwei and Gharbi, Micha{\"e}l and Zhang, Richard and Shechtman, Eli and Durand, Fredo and Freeman, William T and Park, Taesung},
  booktitle={Proceedings of the IEEE/CVF Conference on Computer Vision and Pattern Recognition},
  pages={6613--6623},
  year={2024}
}

@inproceedings{SD_Turbo,
  title={Fast high-resolution image synthesis with latent adversarial diffusion distillation},
  author={Sauer, Axel and Boesel, Frederic and Dockhorn, Tim and Blattmann, Andreas and Esser, Patrick and Rombach, Robin},
  booktitle={SIGGRAPH Asia 2024 Conference Papers},
  pages={1--11},
  year={2024}
}

@article{song2024sdxsrealtimeonesteplatent,
  title={{SDXS}: Real-Time One-Step Latent Diffusion Models with Image Conditions},
  author={Song, Yuda and Sun, Zehao and Yin, Xuanwu},
  journal={arXiv preprint arXiv:2403.16627},
  year={2024}
}

@inproceedings{bolya2023tokenmergingfaststable,
  title={Token merging for fast stable diffusion},
  author={Bolya, Daniel and Hoffman, Judy},
  booktitle={Proceedings of the IEEE/CVF Conference on Computer Vision and Pattern Recognition},
  pages={4599--4603},
  year={2023}
}

@inproceedings{kodaira2025streamdiffusion,
  title={Streamdiffusion: A pipeline-level solution for real-time interactive generation},
  author={Kodaira, Akio and Xu, Chenfeng and Hazama, Toshiki and Yoshimoto, Takanori and Ohno, Kohei and Mitsuhori, Shogo and Sugano, Soichi and Cho, Hanying and Liu, Zhijian and Tomizuka, Masayoshi and others},
  booktitle={2025 IEEE/CVF International Conference on Computer Vision (ICCV)},
  pages={12371--12380},
  year={2025},
  organization={IEEE}
}

@inproceedings{munir2026smoothdiffusion,
  title={SmoothDiffusion-VE: Real-time Generative Video Editing Using Adaptive Feature Cache},
  author={Munir, Mustafa and Zalewski, Sophia and Liu, Shiqiu and Tarjan, David and Belede, Sushmitha and Patney, Anjul and Marculescu, Radu},
  booktitle={2026 IEEE/CVF Winter Conference on Applications of Computer Vision (WACV)},
  pages={8468--8478},
  year={2026},
  organization={IEEE}
}

@inproceedings{liang2025looking,
  title={Looking backward: Streaming video-to-video translation with feature banks},
  author={Liang, Feng and Kodaira, Akio and Xu, Chenfeng and Tomizuka, Masayoshi and Keutzer, Kurt and Marculescu, Diana},
  booktitle={International Conference on Learning Representations},
  volume={2025},
  pages={46425--46445},
  year={2025}
}

@inproceedings{podell2024sdxl,
  title={Sdxl: Improving latent diffusion models for high-resolution image synthesis},
  author={Podell, Dustin and English, Zion and Lacey, Kyle and Blattmann, Andreas and Dockhorn, Tim and M{\"u}ller, Jonas and Penna, Joe and Rombach, Robin},
  booktitle={International Conference on Learning Representations},
  volume={2024},
  pages={1862--1874},
  year={2024}
}

@inproceedings{caron2021emerging,
  title={Emerging properties in self-supervised vision transformers},
  author={Caron, Mathilde and Touvron, Hugo and Misra, Ishan and J{\'e}gou, Herv{\'e} and Mairal, Julien and Bojanowski, Piotr and Joulin, Armand},
  booktitle={Proceedings of the IEEE/CVF International Conference on Computer Vision},
  pages={9650--9660},
  year={2021}
}

@inproceedings{gandikota2026distilling,
  title={Distilling diversity and control in diffusion models},
  author={Gandikota, Rohit and Bau, David},
  booktitle={2026 IEEE/CVF Winter Conference on Applications of Computer Vision (WACV)},
  pages={1304--1313},
  year={2026},
  organization={IEEE}
}

@article{zhao2023unipc,
  title={Unipc: A unified predictor-corrector framework for fast sampling of diffusion models},
  author={Zhao, Wenliang and Bai, Lujia and Rao, Yongming and Zhou, Jie and Lu, Jiwen},
  journal={Advances in Neural Information Processing Systems},
  volume={36},
  pages={49842--49869},
  year={2023}
}

@inproceedings{jena2025elucidating,
  title={Elucidating optimal reward-diversity tradeoffs in text-to-image diffusion models},
  author={Jena, Rohit and Taghibakhshi, Ali and Jain, Sahil and Shen, Gerald and Tajbakhsh, Nima and Vahdat, Arash},
  booktitle={2025 IEEE/CVF Winter Conference on Applications of Computer Vision (WACV)},
  pages={232--242},
  year={2025},
  organization={IEEE}
}

@inproceedings{pan2025tstitch,
  title={{T-Stitch}: Accelerating Sampling in Pre-Trained Diffusion Models with Trajectory Stitching},
  author={Pan, Zizheng and Zhuang, Bohan and Huang, De-An and Nie, Weili and Yu, Zhiding and Xiao, Chaowei and Cai, Jianfei and Anandkumar, Anima},
  booktitle={International Conference on Learning Representations},
  pages={6103--6137},
  year={2025}
}

@inproceedings{taghibakhshi2026star,
title={Star Elastic: Many-in-One Reasoning {LLM}s with Efficient Budget Control},
author={Ali Taghibakhshi and Ruisi Cai and Saurav Muralidharan and Sharath Turuvekere Sreenivas and Ameya Sunil Mahabaleshwarkar and Marcin Chochowski and Akhiad Bercovich and Ran Zilberstein and Ran El-Yaniv and Yonatan Geifman and Daniel Korzekwa and Yoshi Suhara and Oluwatobi Olabiyi and Ashwath Aithal and Nima Tajbakhsh and Pavlo Molchanov},
booktitle={Forty-third International Conference on Machine Learning},
year={2026},
url={https://openreview.net/forum?id=n1fQYnj30I}
}

@inproceedings{munir2025data,
  title={From Data to Design: Leveraging Frequency Statistics for Efficient Neural Network Architectures},
  author={Munir, Mustafa and Li, Guihong and Rahman, Md Mostafijur and Zhang, Alex and Marculescu, Radu},
  booktitle={2025 IEEE/CVF Conference on Computer Vision and Pattern Recognition Workshops (CVPRW)},
  pages={3199--3209},
  year={2025},
  organization={IEEE}
}

@article{cheng2025srdiffusion,
  title={SRDiffusion: Accelerate Video Diffusion Inference via Sketching-Rendering Cooperation},
  author={Cheng, Shenggan and Wei, Yuanxin and Diao, Lansong and Liu, Yong and Chen, Bujiao and Huang, Lianghua and Liu, Yu and Yu, Wenyuan and Du, Jiangsu and Lin, Wei and You, Yang},
  journal={arXiv preprint arXiv:2505.19151},
  year={2025}
}

@inproceedings{song2021denoising,
  title={Denoising Diffusion Implicit Models},
  author={Song, Jiaming and Meng, Chenlin and Ermon, Stefano},
  booktitle={International Conference on Learning Representations},
  year={2021}
}

@inproceedings{blattmann2023stable,
  title={Stable Video Diffusion: Scaling Latent Video Diffusion Models to Large Datasets},
  author={Blattmann, Andreas and Dockhorn, Tim and Kulal, Sumith and Mendelevitch, Daniel and Kilian, Maciej and Lorenz, Dominik and Levi, Yam and English, Zion and Voleti, Vikram and Letts, Adam and others},
  booktitle={International Conference on Learning Representations},
  year={2024}
}

@inproceedings{lu2022dpm,
  title={{DPM-Solver}: A Fast {ODE} Solver for Diffusion Probabilistic Model Sampling in Around 10 Steps},
  author={Lu, Cheng and Zhou, Yuhao and Bao, Fan and Chen, Jianfei and Li, Chunxiao and Zhu, Jun},
  booktitle={Advances in Neural Information Processing Systems},
  year={2022}
}

@inproceedings{ma2024deepcache,
  title={{DeepCache}: Accelerating Diffusion Models for Free},
  author={Ma, Xinyin and Fang, Gongfan and Wang, Xinchao},
  booktitle={Proceedings of the IEEE/CVF Conference on Computer Vision and Pattern Recognition},
  year={2024}
}

@inproceedings{zhao2024pab,
  title={Real-Time Video Generation with Pyramid Attention Broadcast},
  author={Zhao, Xuanlei and Jin, Xiaolong and You, Yang},
  booktitle={Advances in Neural Information Processing Systems},
  year={2024}
}

@inproceedings{luo2023latent,
  title={Latent Consistency Models: Synthesizing High-Resolution Images with Few-Step Inference},
  author={Luo, Simian and Tan, Yiqin and Huang, Longbo and Li, Jian and Zhao, Hang},
  booktitle={International Conference on Learning Representations},
  year={2024}
}

@article{yang2024cogvideox,
  title={{CogVideoX}: Text-to-Video Diffusion Models with An Expert Transformer},
  author={Yang, Zhuoyi and Teng, Jiayan and Zheng, Wendi and Ding, Ming and Huang, Shiyu and Xu, Jiazheng and Yang, Yuanming and Hong, Wenyi and Zhang, Xiaohan and Feng, Guanyu and others},
  journal={arXiv preprint arXiv:2408.06072},
  year={2024}
}

@book{hyvarinen2009natural,
  title={Natural Image Statistics: A Probabilistic Approach to Early Computational Vision},
  author={Hyv{\"a}rinen, Aapo and Hurri, Jarmo and Hoyer, Patrick O.},
  publisher={Springer},
  year={2009}
}

@inproceedings{park2022vision,
  title={How Do Vision Transformers Work?},
  author={Park, Namuk and Kim, Songkuk},
  booktitle={International Conference on Learning Representations},
  year={2022}
}

@inproceedings{rahaman2019spectral,
  title={On the Spectral Bias of Neural Networks},
  author={Rahaman, Nasim and Baratin, Aristide and Arpit, Devansh and Draxler, Felix and Lin, Min and Hamprecht, Fred A. and Bengio, Yoshua and Courville, Aaron},
  booktitle={Proceedings of the 36th International Conference on Machine Learning},
  volume={97},
  pages={5301--5310},
  year={2019},
  publisher={PMLR}
}

@inproceedings{wang2020highfrequency,
  title={High-Frequency Component Helps Explain the Generalization of Convolutional Neural Networks},
  author={Wang, Haohan and Wu, Xindi and Huang, Zeyi and Xing, Eric P.},
  booktitle={Proceedings of the IEEE/CVF Conference on Computer Vision and Pattern Recognition},
  pages={8684--8694},
  year={2020}
}

@article{wan2025,
  title={Wan: Open and Advanced Large-Scale Video Generative Models},
  author={{Team Wan} and Wang, Ang and Ai, Baole and Wen, Bin and Mao, Chaojie and Xie, Chen-Wei and Chen, Di and Yu, Feiwu and Zhao, Haiming and Yang, Jianxiao and Zeng, Jianyuan and others},
  journal={arXiv preprint arXiv:2503.20314},
  year={2025}
}

@article{nvidia2026cosmos,
  title={{Cosmos 3}: Omnimodal World Models for Physical {AI}},
  author={Aditi and Agarwal, Niket and Ali, Arslan and Allen, Jon and Antolini, Martin and Aubame, Adeline and Azzolini, Alisson and Bai, Junjie and Bala, Maciej and Balaji, Yogesh and Bapst, Josh and others},
  journal={arXiv preprint arXiv:2606.02800},
  year={2026}
}

@inproceedings{zhang2025fastvideo,
  title={Fast Video Generation with Sliding Tile Attention},
  author={Zhang, Peiyuan and Chen, Yongqi and Su, Runlong and Ding, Hangliang and Stoica, Ion and Liu, Zhengzhong and Zhang, Hao},
  booktitle={Proceedings of the 42nd International Conference on Machine Learning},
  year={2025}
}

@article{zhang2025turbodiffusion,
  title={Turbodiffusion: Accelerating video diffusion models by 100-200 times},
  author={Zhang, Jintao and Zheng, Kaiwen and Jiang, Kai and Wang, Haoxu and Stoica, Ion and Gonzalez, Joseph E and Chen, Jianfei and Zhu, Jun},
  journal={arXiv preprint arXiv:2512.16093},
  year={2025}
}

@article{balaji2022ediff,
  title={ediff-i: Text-to-image diffusion models with an ensemble of expert denoisers},
  author={Balaji, Yogesh and Nah, Seungjun and Huang, Xun and Vahdat, Arash and Song, Jiaming and Zhang, Qinsheng and Kreis, Karsten and Aittala, Miika and Aila, Timo and Laine, Samuli and others},
  journal={arXiv preprint arXiv:2211.01324},
  year={2022}
}

@inproceedings{pan2023stitchable,
  title={Stitchable neural networks},
  author={Pan, Zizheng and Cai, Jianfei and Zhuang, Bohan},
  booktitle={2023 IEEE/CVF Conference on Computer Vision and Pattern Recognition (CVPR)},
  pages={16102--16112},
  year={2023},
  organization={IEEE}
}

@inproceedings{pan2024snnetv2,
  title={Stitched vits are flexible vision backbones},
  author={Pan, Zizheng and Liu, Jing and He, Haoyu and Cai, Jianfei and Zhuang, Bohan},
  booktitle={European Conference on Computer Vision},
  pages={258--274},
  year={2024},
  organization={Springer}
}

@inproceedings{yang2023diffusion,
  title={Diffusion probabilistic model made slim},
  author={Yang, Xingyi and Zhou, Daquan and Feng, Jiashi and Wang, Xinchao},
  booktitle={2023 IEEE/CVF Conference on Computer Vision and Pattern Recognition (CVPR)},
  pages={22552--22562},
  year={2023},
  organization={IEEE}
}

@InProceedings{Sun_t2v_compbench_2025_CVPR,
    author    = {Sun, Kaiyue and Huang, Kaiyi and Liu, Xian and Wu, Yue and Xu, Zihan and Li, Zhenguo and Liu, Xihui},
    title     = {T2V-CompBench: A Comprehensive Benchmark for Compositional Text-to-video Generation},
    booktitle = {Proceedings of the IEEE/CVF Conference on Computer Vision and Pattern Recognition (CVPR)},
    month     = {June},
    year      = {2025},
    pages     = {8406-8416}
}

@InProceedings{Liu_evalcrafter_2024_CVPR,
    author    = {Liu, Yaofang and Cun, Xiaodong and Liu, Xuebo and Wang, Xintao and Zhang, Yong and Chen, Haoxin and Liu, Yang and Zeng, Tieyong and Chan, Raymond and Shan, Ying},
    title     = {EvalCrafter: Benchmarking and Evaluating Large Video Generation Models},
    booktitle = {Proceedings of the IEEE/CVF Conference on Computer Vision and Pattern Recognition (CVPR)},
    month     = {June},
    year      = {2024},
    pages     = {22139-22149}
}

@inproceedings{fu2023dreamsim,
  title={DreamSim: Learning New Dimensions of Human Visual Similarity using Synthetic Data},
  author={Fu, Stephanie and Tamir, Netanel and Sundaram, Shobhita and Chai, Lucy and Zhang, Richard and Dekel, Tali and Isola, Phillip},
  booktitle={Advances in Neural Information Processing Systems},
  year={2023}
}
